\RequirePackage{fix-cm}
\documentclass{externalized}
\usepackage{microtype}
\usepackage{graphicx}
\usepackage{hyperref}
\usepackage{url}
\usepackage{booktabs}
\usepackage{amsfonts}
\usepackage{amsmath}
\usepackage{amssymb}
\usepackage{amsthm}
\usepackage{mathtools}
\usepackage{enumitem}
\usepackage{multirow}
\usepackage{subcaption}
\usepackage{xcolor}
\usepackage{listings}
\usepackage{natbib}
\usepackage{multicol}
\usepackage{pifont}
\usepackage{xspace}
\usepackage{wrapfig}
\usepackage{float}
\usepackage{tabularx}
\usepackage{ragged2e}
\usepackage{longtable}
\usepackage{colortbl}

\DeclareTextFontCommand{\tabletextbf}{\fontseries{bx}\selectfont}
\AtBeginEnvironment{tabular}{\let\textbf\tabletextbf}
\AtBeginEnvironment{tabular*}{\let\textbf\tabletextbf}
\AtBeginEnvironment{tabularx}{\let\textbf\tabletextbf}
\AtBeginEnvironment{longtable}{\let\textbf\tabletextbf}

\providecommand{\noopsort}[1]{}
\AtBeginDocument{%
  \providecommand\BibTeX{{%
    \normalfont B\kern-0.5em{\scshape i\kern-0.25em b}\kern-0.8em\TeX}}}

\makeatletter
\DeclareRobustCommand\onedot{\futurelet\@let@token\@onedot}
\def\@onedot{\ifx\@let@token.\else.\null\fi}

\makeatother

\definecolor{codegreen}{rgb}{0,0.6,0}
\definecolor{codegray}{rgb}{0.5,0.5,0.5}
\definecolor{codepurple}{rgb}{0.58,0,0.82}
\definecolor{backcolour}{rgb}{0.95,0.95,0.92}

\lstdefinestyle{mystyle}{
  backgroundcolor=\color{backcolour},
  commentstyle=\color{codegreen},
  keywordstyle=\color{magenta},
  numberstyle=\tiny\color{codegray},
  stringstyle=\color{codepurple},
  basicstyle=\ttfamily\footnotesize,
  breakatwhitespace=false,
  breaklines=true,
  captionpos=b,
  keepspaces=true,
  numbers=left,
  numbersep=5pt,
  showspaces=false,
  showstringspaces=false,
  showtabs=false,
  tabsize=2
}

\usepackage{needspace}
\usepackage{fontawesome5}
\usepackage{xurl}
\usepackage{tikz}
\usetikzlibrary{arrows.meta,positioning,calc,shapes.geometric}

\newcommand{\TableOuterRules}[1]{%
  \leavevmode\vbox{%
    \hrule height 1.5pt
    \hbox{\ignorespaces#1\unskip}%
    \hrule height 1.5pt
  }%
}
\newcommand{\MaskedFigure}[2][]{\includegraphics[#1]{#2}}

\newcommand{\AffiliationOutlineCircle}{%
  (0.5,0.5) circle[radius=0.5]}

\newcommand{\AffiliationOutlineQiji}{%
  [rounded corners=1.2mm] (0,0) rectangle (1,1)}

\newcommand{\AffiliationOutlineSiiMark}{%
  (0.195946,0.689189) --
  (0.209459,0.754054) --
  (0.222973,0.770270) --
  (0.260811,0.789189) --
  (0.314865,0.797297) --
  (0.550000,0.802703) --
  (0.585135,0.843243) --
  (0.641892,0.883784) --
  (0.690541,0.937838) --
  (0.701351,0.937838) --
  (0.709459,0.954054) --
  (0.739189,0.972973) --
  (0.750000,0.991892) --
  (0.766216,0.986486) --
  (0.760811,0.932432) --
  (0.768919,0.902703) --
  (0.814865,0.835135) --
  (0.817568,0.810811) --
  (0.814865,0.789189) --
  (0.795946,0.764865) --
  (0.587838,0.594595) --
  (0.704054,0.535135) --
  (0.731081,0.508108) --
  (0.731081,0.481081) --
  (0.658108,0.410811) --
  (0.250000,0.083784) --
  (0.217568,0.045946) --
  (0.201351,0.002703) --
  (0.195946,0.235135) --
  (0.206757,0.289189) --
  (0.236486,0.321622) --
  (0.409459,0.459459) --
  (0.417568,0.475676) --
  (0.214865,0.567568) --
  (0.198649,0.597297) -- cycle}

\newsavebox{\AffiliationLogoBox}
\newcommand{\instlogoheight}{12mm}
\newcommand{\instlogogap}{5mm}
\newcommand{\instlogoraise}{2mm}
\newcommand{\AffiliationLogoShadow}[1]{%
  \begin{scope}[xshift=0.85pt,yshift=-0.95pt,
      transparency group,opacity=0.32]
    \path[fill=black] #1;
    \foreach \ShadowWidth/\ShadowOpacity in
        {2.4/0.015,2.0/0.035,1.6/0.055,1.2/0.095,0.8/0.15,0.4/0.25}{%
      \path[draw=black,line width=\ShadowWidth pt,
        opacity=\ShadowOpacity,line join=round] #1;
    }
  \end{scope}%
}
\DeclareRobustCommand{\AffiliationLogo}[3][]{%
  \sbox{\AffiliationLogoBox}{\includegraphics[#1,
    width=\instlogoheight,height=\instlogoheight,clip]{#3}}%
  \raisebox{-0.5\height}{\begin{tikzpicture}[baseline=0pt,
      x=\wd\AffiliationLogoBox,y=\ht\AffiliationLogoBox]
    \path[use as bounding box] (0,0) rectangle (1,1);
    \begin{scope}[overlay,yshift=\instlogoraise]
      \AffiliationLogoShadow{#2}%
      \path[fill=white] #2;
      \begin{scope}
        \path[clip] #2;
        \node[anchor=south west,inner sep=0pt,outer sep=0pt] at (0,0)
          {\usebox{\AffiliationLogoBox}};
      \end{scope}
    \end{scope}
  \end{tikzpicture}}%
}

\DeclareRobustCommand{\AffiliationStandaloneLogo}[1]{%
  \raisebox{-0.5\height}{\begin{tikzpicture}[baseline=0pt,
      x=\instlogoheight,y=\instlogoheight]
    \path[use as bounding box] (0,0) rectangle (1,1);
    \begin{scope}[overlay,yshift=\instlogoraise]
      \AffiliationLogoShadow{\AffiliationOutlineSiiMark}%
      \node[anchor=south,inner sep=0pt,outer sep=0pt] at (0.5,0)
        {\includegraphics[height=\instlogoheight,keepaspectratio]{#1}};
    \end{scope}
  \end{tikzpicture}}%
}

\newcommand{\WorkRoverTitle}{%
  \raisebox{-0.35em}{\includegraphics[height=1.5em,viewport=200 211 1916 563,clip]{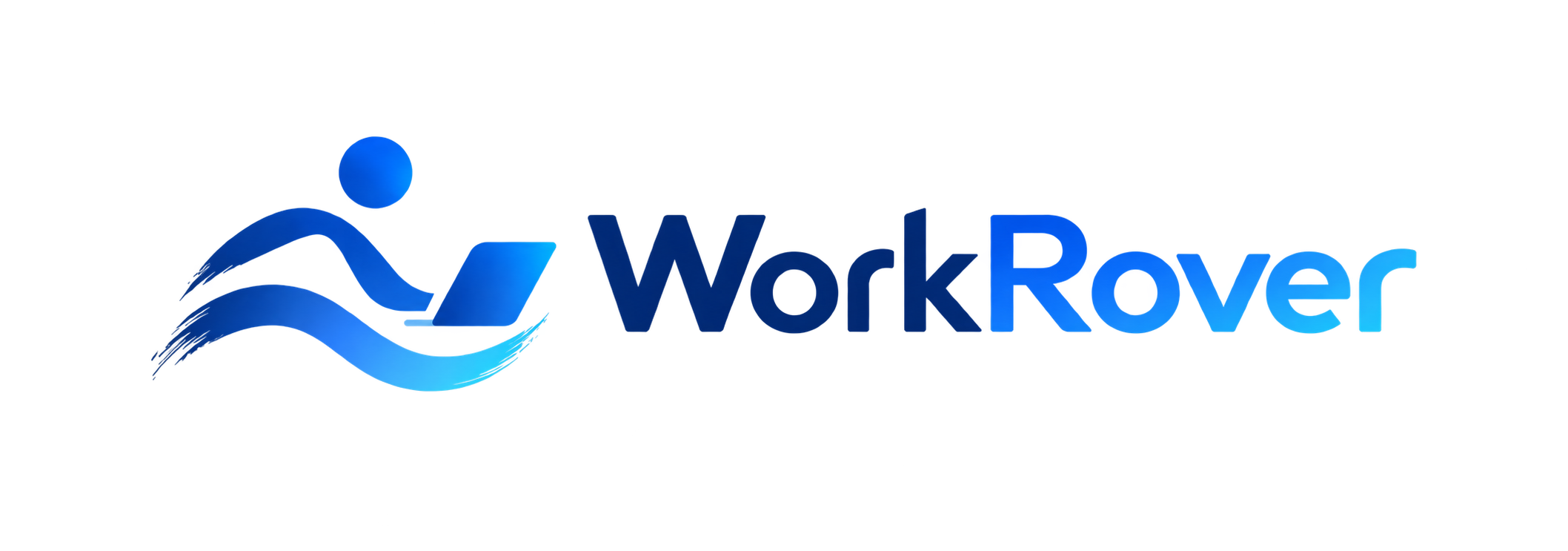}}}
\title{\texorpdfstring{\WorkRoverTitle}{WorkRover}: Scaling Long-Horizon Cowork Agent Training Via Recursive Environment Evolution}

\renewcommand{\institutionlogosleft}{%
  \AffiliationLogo{\AffiliationOutlineQiji}{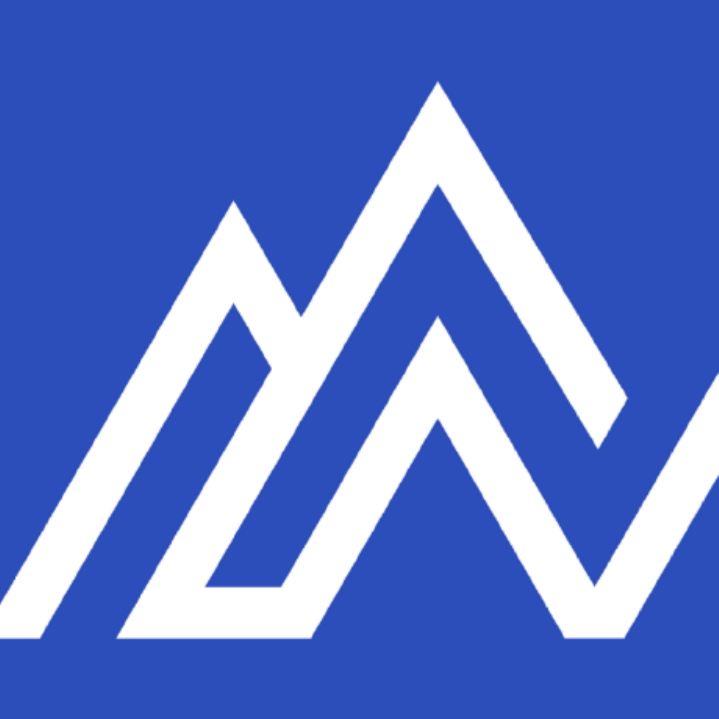}%
  \hspace{\instlogogap}%
  \AffiliationLogo[trim=39bp 38bp 39bp 40bp]{\AffiliationOutlineCircle}{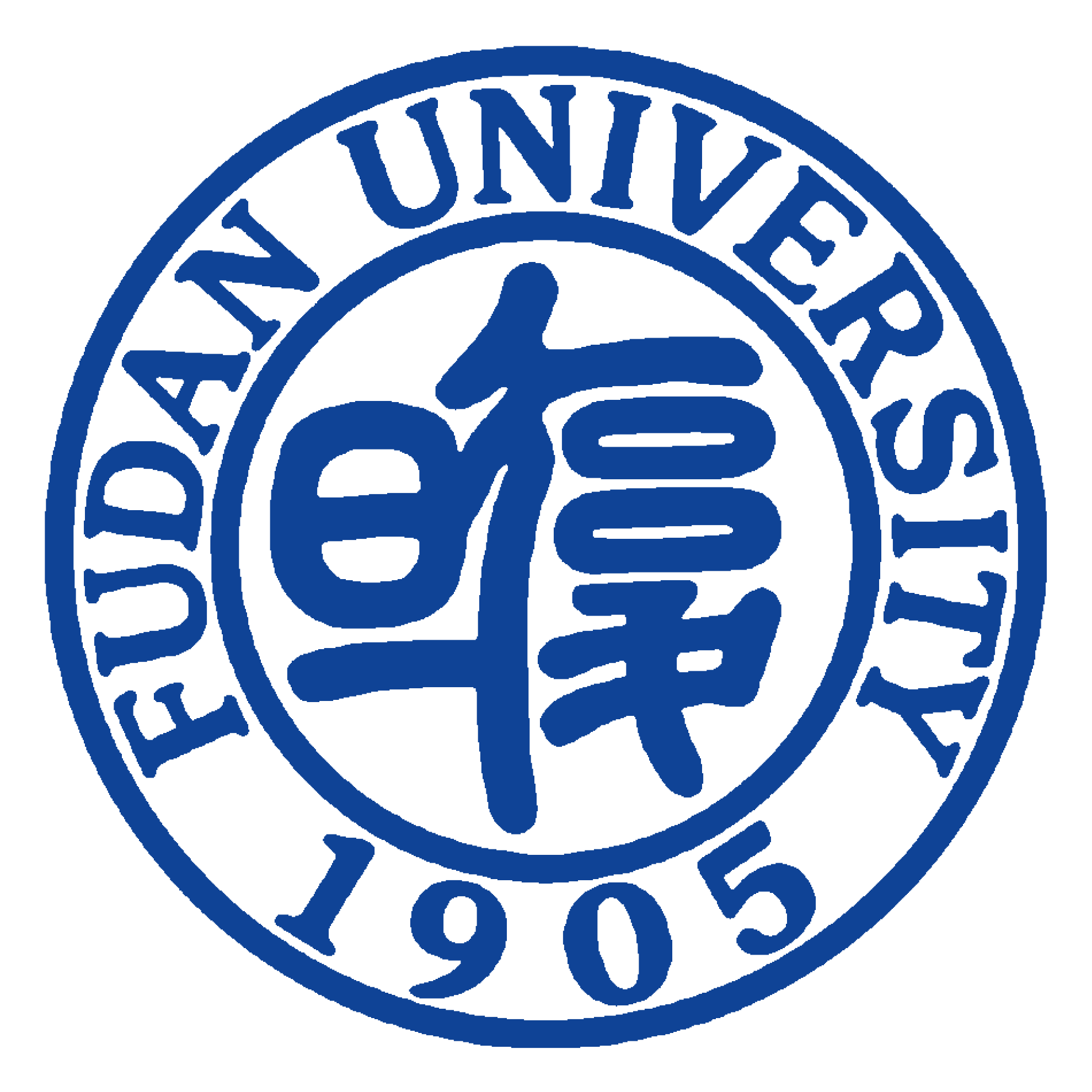}%
  \hspace{\instlogogap}%
  \AffiliationLogo[trim=10bp 10bp 12bp 12bp]{\AffiliationOutlineCircle}{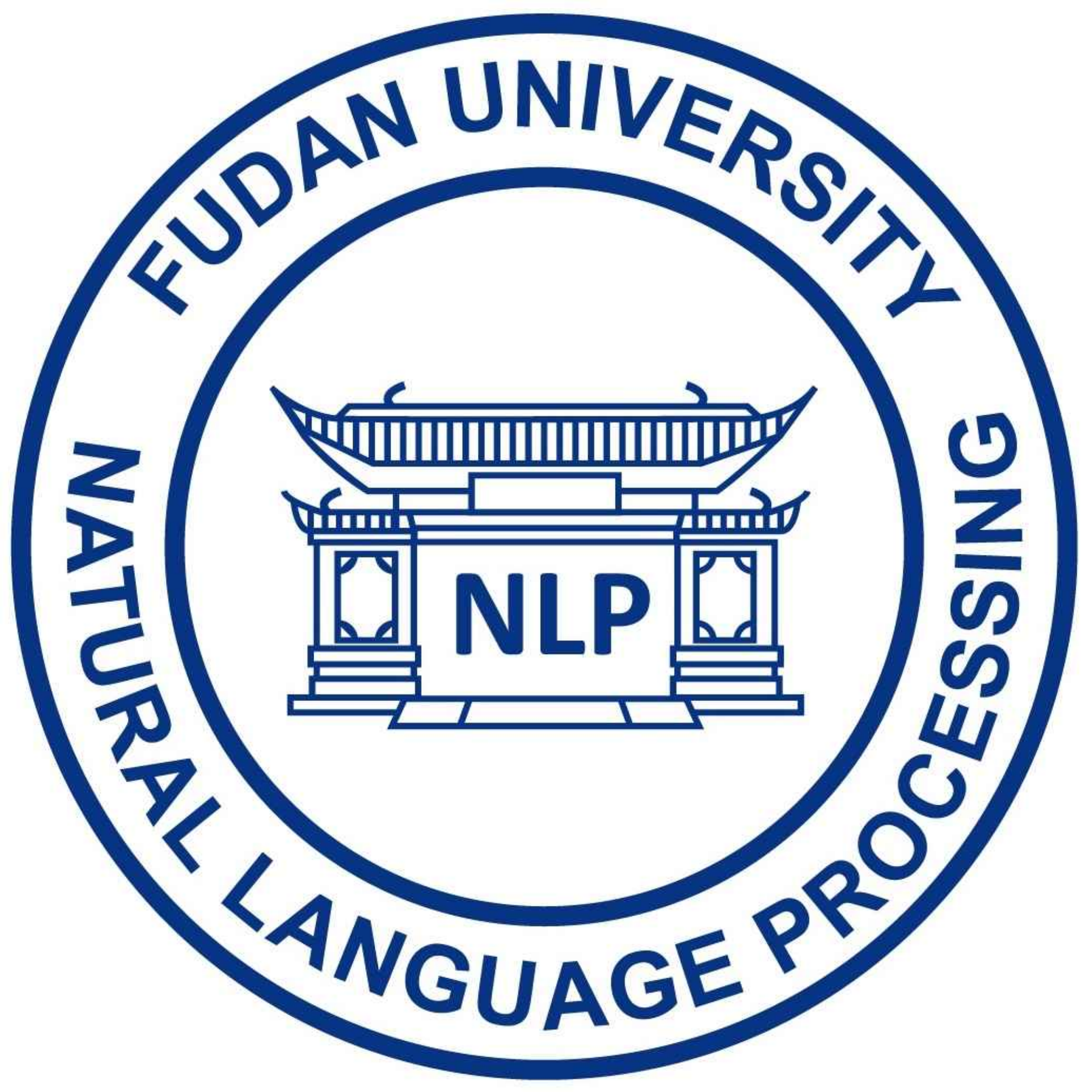}%
}
\renewcommand{\institutionlogosright}{%
  \AffiliationLogo[trim=1bp 1bp 1bp 1bp]{\AffiliationOutlineCircle}{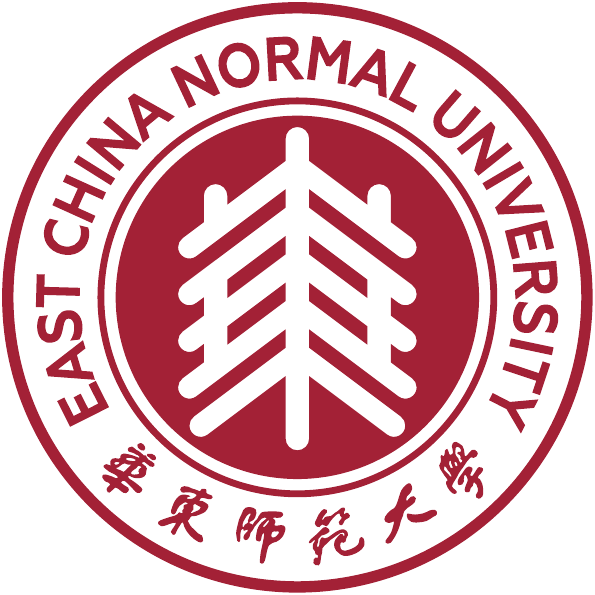}%
  \hspace{\instlogogap}%
  \AffiliationLogo[trim=1bp 1bp 1bp 1bp]{\AffiliationOutlineCircle}{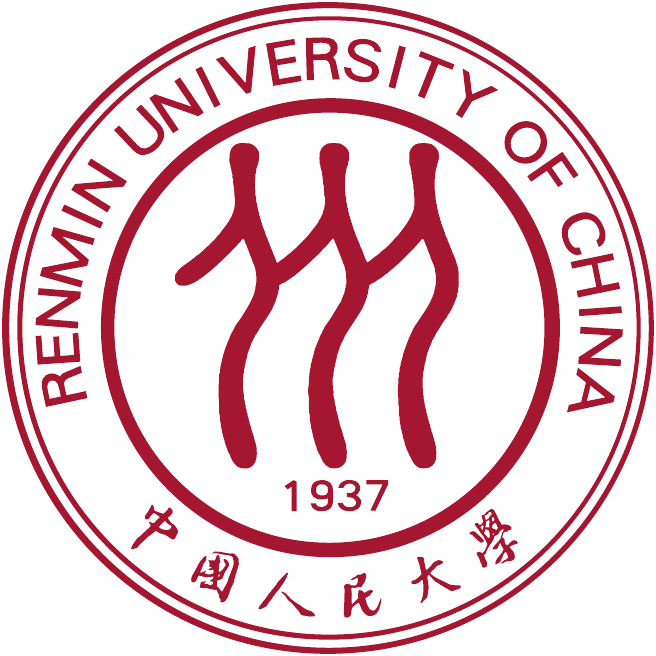}%
  \hspace{\instlogogap}%
  \AffiliationStandaloneLogo{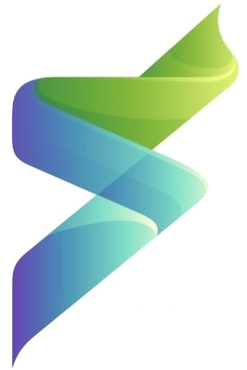}%
}

\author[*]{\mbox{Hang He}}
\author[*]{\mbox{Linting Wang}}
\author[*]{\mbox{Guanming Liu}}
\author[*]{\mbox{Bo Mao}}
\author[*]{\mbox{Jinxiu Liu}}
\author[*]{\mbox{Yiran Suo}}
\author{\\\mbox{Maosen Zhou}}
\author{\mbox{Zhiheng Xi}}
\author{\mbox{Guanting Dong}}
\author[\;\text{\faIcon[regular]{envelope}}]{\mbox{Zhou Shao}}
\author{\mbox{Rui Zheng}}
\author{\mbox{Jie Zhou}}
\author{\\\mbox{Chengcheng Wan}}
\author[\;\text{\faIcon[regular]{envelope}}]{\mbox{Tao Gui}}
\author{\mbox{Qi Zhang}}
\author{\mbox{Xipeng Qiu}}
\author{\mbox{Xuanjing Huang}}

\contribution[*]{Equal Contribution}
\contribution[\text{\faIcon[regular]{envelope}}]{Corresponding Authors}
\affiliation{Fudan University, East China Normal University, Renmin University of China,\\
Shanghai Qiji Zhifeng Co., Ltd., Shanghai Innovation Institute}

\abstract{%
Scaling interactive environments has become increasingly important for training long-horizon agents through reinforcement learning.
These environments often require further testing and refinement after initial generation, and correcting defects in a task may also require updating its work context and verifier.
However, existing construction pipelines offer limited support for such iterative refinement while keeping these components aligned.
In this paper, we introduce \textbf{\textsc{WorkRover}}, a framework for training long-horizon cowork agents through recursive environment evolution.
Its harness-driven synthesis enables construction agents to jointly build and validate semantically aligned work contexts, executable tasks, and evidence-grounded verifiers, adapting their actions to workspace state and execution feedback rather than following a fixed construction workflow.
To support environment evolution, we introduce \textbf{Factorial Compatibility Replay (FCR)}, which uses rollout feedback to evaluate combinations of component revisions through dependency-closed replay and selects the smallest validated repair among tested candidates while preserving task requirements and evaluation constraints.
To provide finer-grained training signals, we propose \textbf{Verifier-to-Turn Credit Assignment (VTCA)}, which augments trajectory-level GRPO advantages with signals from deterministic workspace-progress checks and assigns a bounded share of positive-progress credit to prerequisite turns using recorded tool-output and artifact dependencies.
Using \textsc{WorkRover}, we post-train Qwen3.5-35B-A3B to obtain \textsc{WorkRover}-35B-A3B and evaluate it on several cowork-agent benchmarks.
Its reported scores exceed those of Qwen3.5-35B-A3B by 26.27 percentage points on JobBench, 29.32 percentage points on OfficeQA Pro, and 413.25 Elo points on GDPval v2.
These results suggest that \textsc{WorkRover} offers a promising foundation for scalable training of long-horizon cowork agents.
\par\smallskip
{\centering\href{https://workrover.github.io/}{\textcolor{themeaccent}{\faIcon{globe}}}\hspace{0.4em}\textbf{Project Page:} \url{https://workrover.github.io/}\par}
}

\renewcommand{\topfraction}{0.9}
\renewcommand{\bottomfraction}{0.85}
\renewcommand{\textfraction}{0.08}
\renewcommand{\floatpagefraction}{0.85}

\begin{document}
\newgeometry{top=2cm,bottom=2.5cm,left=2.5cm,right=2.5cm,
  headheight=52pt,headsep=-2.5mm}
\raggedbottom
\clubpenalties 2 10000 0
\maketitle
\enlargethispage{1.5\baselineskip}
\setlength{\parskip}{4pt plus 1pt minus 1pt}

\begingroup
\setlength{\intextsep}{3pt}
\begin{figure}[H]
    \centering
    \setlength{\abovecaptionskip}{4pt}
    \captionsetup{width=\dimexpr\linewidth-2\leftmargini\relax,font=small}
    \vbox{\offinterlineskip
        \hbox to \linewidth{\hfil
            \MaskedFigure[viewport=12 246 710 406,clip,width=0.84\linewidth]{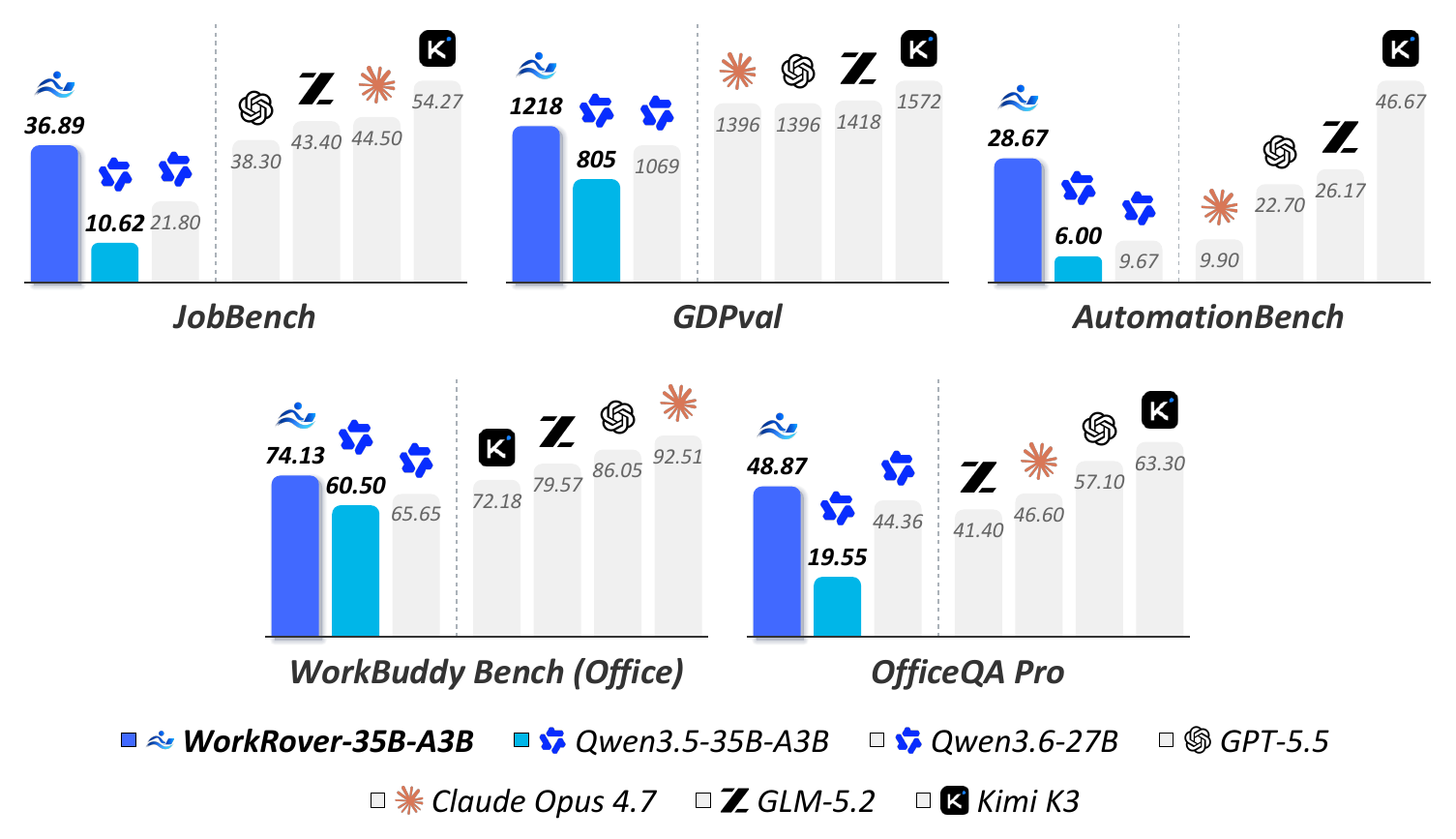}\hfil}%
        \kern4pt
        \hbox to \linewidth{\hfil
            \MaskedFigure[viewport=12 70 710 230,clip,width=0.84\linewidth]{figures/generated/figure1c_five_charts.pdf}\hfil}%
        \kern3pt
        \hbox to \linewidth{\hfil\resizebox{\linewidth}{!}{%
            \hbox to 920bp{\hfil
                \MaskedFigure[viewport=59 37 237 61,clip,height=22bp]{figures/generated/figure1c_five_charts.pdf}\hfil
                \MaskedFigure[viewport=254 37 413 61,clip,height=22bp]{figures/generated/figure1c_five_charts.pdf}\hfil
                \MaskedFigure[viewport=430 37 557 61,clip,height=22bp]{figures/generated/figure1c_five_charts.pdf}\hfil
                \MaskedFigure[viewport=574 37 662 61,clip,height=22bp]{figures/generated/figure1c_five_charts.pdf}\hfil
                \MaskedFigure[viewport=182 7 328 31,clip,height=22bp]{figures/generated/figure1c_five_charts.pdf}\hfil
                \MaskedFigure[viewport=344 7 437 31,clip,height=22bp]{figures/generated/figure1c_five_charts.pdf}\hfil
                \MaskedFigure[viewport=453 7 539 31,clip,height=22bp]{figures/generated/figure1c_five_charts.pdf}\hfil
            }%
        }\hfil}%
    }%
    \caption{\textbf{Cross-benchmark performance.} Results on five cowork-agent benchmarks.}
    \label{fig:motivation-performance}
\end{figure}
\endgroup
\restoregeometry

\section{Introduction}
\label{sec:introduction}

Cowork agents powered by large language models (LLMs) perform professional tasks across files, tools, and applications~\citep{li2026scalablellmagenttoolaccess}.
These tasks unfold over long sequences of dependent actions, requiring agents to locate relevant materials~\citep{su2026coalrag,he2026localsearchbenchbenchmarkingagentic,he2026vistahop}, track workspace changes, and maintain consistency across related artifacts \citep{niu2026dumatebench,tang2026workspacebench10benchmarkingai,he2026checkerbench,sun2026brat}.
Training these agents through reinforcement learning (RL) depends on environments that support diverse, executable workflows and provide reliable feedback on task outcomes \citep{hao2026multisteptooluserl,cheng2026efficientreinforcementlearning,chai2025rlfactoryplugandplayreinforcementlearning,mao2026weft,xi2025agentgymrl,zeng2026recrmbench,zhou2026ragstress}.
Recent work has explored automated environment and task synthesis to expand the range of workspaces and tasks available for agent training \citep{song2026envscalerscalingtoolinteractive,bai2026clawgymscalableframeworkbuilding,chen2026toolforge}.


However, expanding the number of synthesized environments does not by itself ensure reliable training feedback.
In a cowork environment, the work context must support the task's requirements, and the verifier must assess the requested outcomes while accepting valid alternative solutions.
Agent rollouts can expose missing dependencies, inconsistent task specifications, or verification errors that remain after initial construction \citep{zhong2026hackerfixer}.
Such defects can make valid behavior appear unsuccessful or allow incorrect results to be rewarded.
Repairing these defects often requires coordinated revisions within a task environment: a correction to the work context or task specification may also require updating the verifier to keep execution and evaluation aligned.
Existing construction pipelines offer limited support for maintaining this alignment throughout repeated revisions.
This motivates \emph{recursive environment evolution}: repeatedly using execution feedback to identify defects, coordinate component revisions, and revalidate the revised environment while preserving the original task requirements and evaluation constraints.
Reliable terminal evaluation also leaves a credit-assignment challenge: useful steps can occur in failed trajectories and redundant actions in successful ones, while earlier information-gathering turns may only contribute through later workspace changes \citep{tao2026trace,dong2026tapotoolawarepolicyoptimization}.

\begin{figure}[!t]
    \centering
    \MaskedFigure[width=\linewidth]{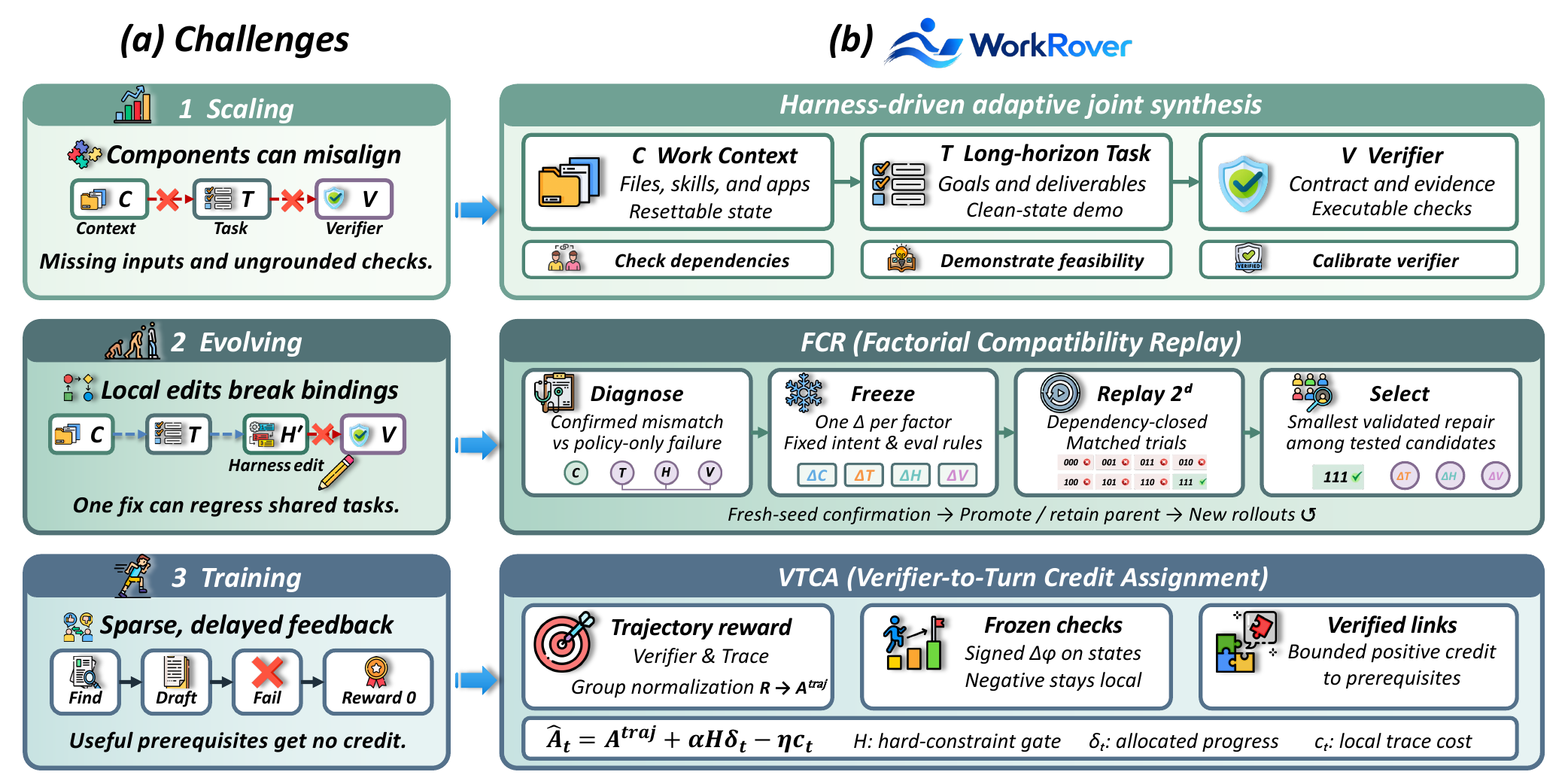}
    \captionsetup{justification=centering,singlelinecheck=false}
    \caption{\textbf{Cowork agents: challenges and approaches.}}
    \label{fig:motivation}
\end{figure}

To this end, we introduce \textbf{\textsc{WorkRover}}, a framework for long-horizon cowork agent training that integrates harness-driven environment synthesis, recursive environment evolution, and verifier-grounded policy learning
(\hyperref[fig:motivation]{Figure~\ref*{fig:motivation}(b)}).
Its harness tracks component dependencies and enables construction agents to jointly build and validate semantically aligned work contexts, executable tasks, and evidence-grounded verifiers.
These agents adapt their construction actions to workspace state and execution feedback rather than following a fixed workflow.
Independent reference executions from a clean initial workspace establish task feasibility, while verifier validation tests incorrect outcomes and valid alternatives against instruction-grounded requirements rather than a prescribed solution path.
For environment evolution, we introduce \emph{Factorial Compatibility Replay} (FCR), which evaluates combinations of fixed repair proposals across the work context, task, harness, and verifier.
Each combination is rebuilt with its dependency-induced updates and tested under matched rollout conditions.
FCR selects the smallest validated repair among the tested combinations, subject to preserving task requirements and evaluation constraints and passing regression checks on affected tasks.
Fresh rollouts confirm selected repairs and provide evidence for subsequent evolution.
For policy learning, we introduce \emph{Verifier-to-Turn Credit Assignment} (VTCA) to supplement trajectory-level GRPO advantages with finer-grained feedback.
Frozen deterministic checks measure progress and setbacks in recorded workspace states, while tool-output and artifact dependencies identify prerequisite turns eligible for a bounded share of positive-progress credit.
This provides turn-level learning signals without intermediate judge calls.

We post-train Qwen3.5-35B-A3B with \textsc{WorkRover} and evaluate the resulting model on five external benchmarks spanning occupational tasks, professional deliverables, cross-application workflows, and document-grounded reasoning.
\textsc{WorkRover}-35B-A3B exceeds the reported backbone scores on all five benchmarks, with gains of 26.27 percentage points on JobBench, 29.32 percentage points on OfficeQA Pro, and 413.25 Elo points on GDPval v2.
Controlled experiments further examine the contributions of environment evolution and verifier-grounded learning.
Under matched repair budget caps, FCR improves task-package synthesis pass rate by 6.0 percentage points over full regeneration.
A separate maintenance evaluation shows that FCR also resolves naturally occurring defects more reliably while reducing unintended changes to task intent.
Construction ablations also show that following file dependencies and matching files with compatible skills improve task yield and executability, beyond assembling workspaces through topical similarity alone.
With the supervised fine-tuning (SFT) initialization, task and seed schedules, and rollout budget fixed, VTCA outperforms episode-only GRPO across all five benchmarks.
These findings support improving both environment reliability and turn-level learning signals for long-horizon cowork agent training.

\Needspace{4\baselineskip}
Our main contributions are:\par\nopagebreak[4]
\begingroup
\setlength{\leftmargini}{1.7em}
\begin{enumerate}
    \setlength{\topsep}{0pt}
    \setlength{\partopsep}{0pt}
    \setlength{\itemsep}{2pt}
    \setlength{\parsep}{0pt}
    \setlength{\parskip}{0pt}
    \interlinepenalty=0
    \item \textbf{Harness-driven construction of semantically aligned environments.} We introduce a Harness-driven synthesis pipeline that jointly constructs \emph{semantically aligned} work contexts, executable tasks, and evidence-grounded verifiers at scale. The Harness explicitly \emph{tracks their dependencies} and validates \emph{task feasibility and verifier consistency}.
    \item \textbf{Recursive environment evolution.}
    We introduce \emph{Factorial Compatibility Replay} (FCR) to support iterative environment repair under explicit component dependencies. FCR evaluates coordinated revisions, identifies \emph{minimal validated repairs}, and propagates changes across affected components while \emph{preserving task intent and evaluation constraints}.
    \item \textbf{Training from verifier-grounded progress.}
    We introduce \emph{Verifier-to-Turn Credit Assignment} (VTCA) to augment
    terminal-reward GRPO with \emph{workspace-grounded advantage shaping}. Deterministic
    checks provide progress proxies, with bounded positive allocations to
    \emph{prerequisite turns} through recorded tool and artifact dependencies,
    without intermediate Judge calls.
\end{enumerate}
\endgroup

\section{Related Work}
\label{sec:related-work}

\paragraph{Executable environments and verification.}
Prior work scales executable environments~\citep{
trivedi2024appworldcontrollableworldapps,song2026envscalerscalingtoolinteractive,
bai2026clawgymscalableframeworkbuilding,dong2026agentworld,xi2025agentgym}
and validated task synthesis~\citep{chen2025toolforgedatasynthesispipeline,chen2026toolforge,
pan2026metataskturningterminaltask,hua2026cliuniverse}.
Shared specifications and state help align tasks and
verifiers~\citep{ivanov2026anchor,shi2026facet}, while adversarial audits test
verifier reliability~\citep{zhong2026hackerfixer}.
\textsc{WorkRover} jointly constructs Work Contexts, Tasks, and Verifiers,
tracking their dependencies under a versioned Harness.

\paragraph{Environment evolution and repair.}
Recent work studies task-bundle revision and execution-driven environment
repair~\citep{wu2026envsforge,pandya2026echoverse,mao2026weft}.
Program repair also studies patch minimization~\citep{weimer2009genprog}.
FCR replays combinations of fixed repair proposals, including dependency-induced
changes. It selects the smallest validated repair among those tested, preserving
task requirements and evaluation constraints and checking affected tasks for
regressions.

\paragraph{Long-horizon learning and progress attribution.}
Local credit assignment draws on state comparisons and planning
hierarchies~\citep{feng2025gigpo,peng2026hiperhierarchicalreinforcementlearning},
progress estimates, or evidence attribution~\citep{
tao2026trace,zhang2026trca,xu2026stamp,gandhi2026draco,yue2026promotingefficientreasoning,xi2025agentprm}.
VTCA shapes GRPO advantages~\citep{shao2024deepseekmath} with frozen checks of
workspace progress and setbacks. It assigns a bounded share of positive credit
to prerequisite turns using recorded tool-output and artifact-version
dependencies, without intermediate Judge calls.
Checker coverage~\citep{zhou2026coveragenottargeting} and dependency
quality limit these signals. See Appendix~\ref{app:related-work}.

\section{Method}
\label{sec:method}

\begin{figure}[!t]
    \centering
    \includegraphics[width=\linewidth]{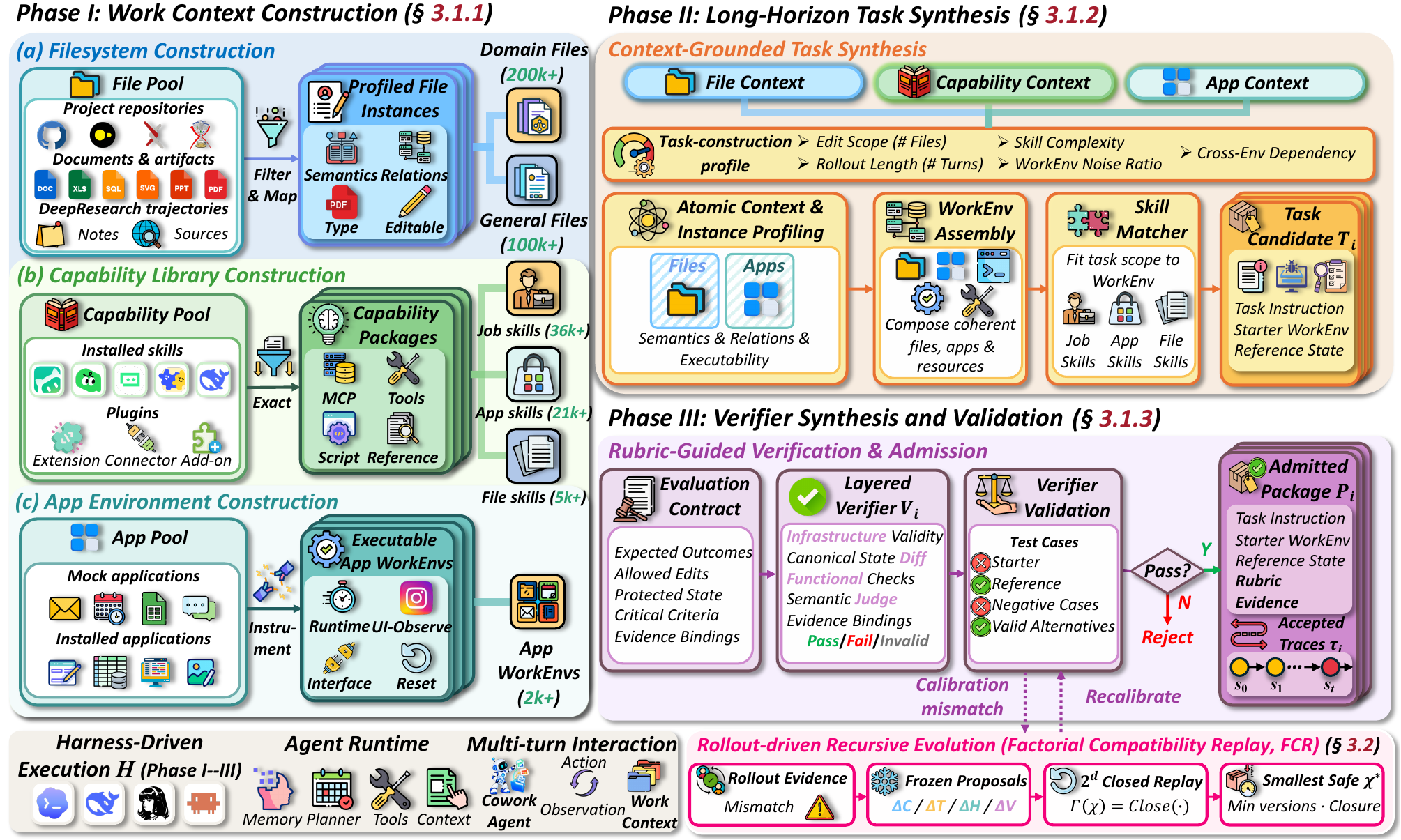}
    \caption{Environment construction and recursive evolution in \textsc{WorkRover}.
    The Harness coordinates Work Context construction, Task synthesis, and
    Verifier synthesis and validation. FCR repairs compatibility mismatches under
    frozen safety gates; verified trajectories support policy learning.}
    \label{fig:WorkRover-overview}
\end{figure}

In this section, we present \textbf{\textsc{WorkRover}}
(Figure~\ref{fig:WorkRover-overview}), a framework for training long-horizon
cowork agents through recursive environment evolution. We first introduce
harness-driven synthesis
of semantically aligned \textit{work contexts}, \textit{tasks}, and \textit{verifiers}
(Sec.~\ref{sec:environment-synthesis}).
We then present \textbf{Factorial Compatibility Replay (FCR)} for rollout-driven
recursive evolution (Sec.~\ref{sec:workspace-coevolution}) and
\textbf{Verifier-to-Turn Credit Assignment (VTCA)} (Sec.~\ref{sec:training}).

\subsection{Harness-Driven Environment Synthesis}
\label{sec:environment-synthesis}

We organize the training environments into $N$ task packages:
{\small
\[
\mathfrak{L}
=
\left\{
\mathcal{P}_i
=
\left\langle C_i,T_i,V_i\right\rangle
\right\}_{i=1}^{N}.
\]
}
The Work Context $C_i$ provides files, capabilities, and application
environments; the Task $T_i$ specifies goals, deliverables, and constraints;
and the Verifier $V_i$ checks task completion. The Harness $H$ coordinates
construction, tracks dependencies, and manages execution across the three
phases in Figure~\ref{fig:WorkRover-overview}.

\subsubsection{Work Context Construction}
\label{sec:work-context-construction}

\begin{figure}[!htbp]
\centering
\begin{minipage}[c]{0.75\linewidth}
\centering
\MaskedFigure[width=0.9\linewidth]{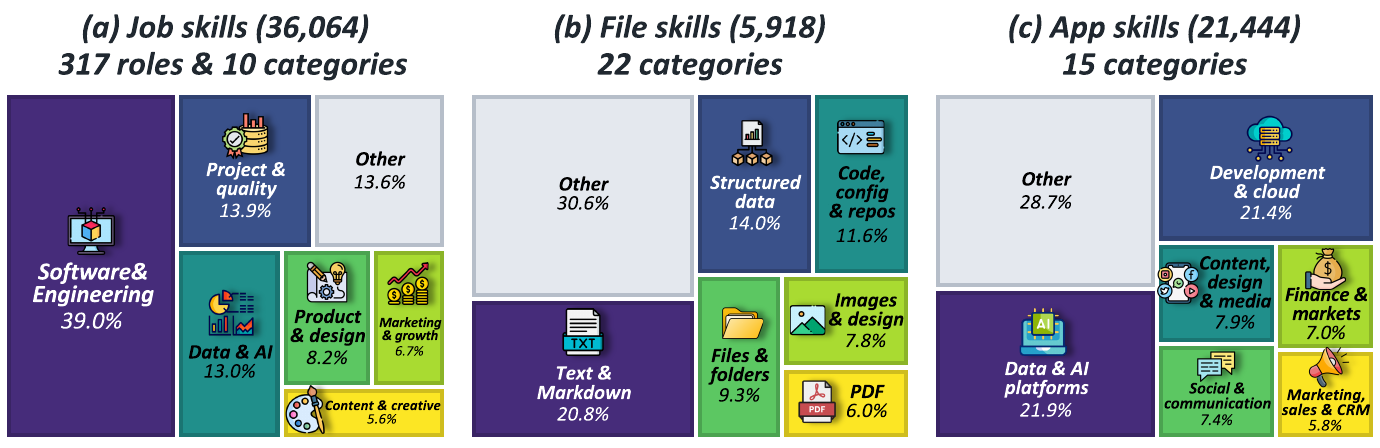}
\end{minipage}\hfill
\begin{minipage}[c]{0.24\linewidth}
\setlength{\abovecaptionskip}{0pt}
\setlength{\belowcaptionskip}{0pt}
\caption{\RaggedRight\textbf{Skill-library coverage.}
Six largest categories; others aggregated.
Tile sizes show shares of skills~(a,~c) or category
\mbox{memberships}~(b).}
\label{fig:skill-distributions}
\end{minipage}
\end{figure}

Phase~I constructs a reusable Work Context $C_i$ comprising File Context
$C_i^{\mathrm{file}}$, Capability Context $C_i^{\mathrm{cap}}$, and App Context
$C_i^{\mathrm{app}}$.
We collect files from project repositories and DeepResearch trajectories,
along with documents and artifacts. After filtering, we profile their
semantics, cross-file relations, type, and editability to form general-purpose
and domain-specific collections.
We package installed skills and plugins with their MCP interfaces, tools,
scripts, and references. These packages form job, file, and app skill libraries.
We instrument mock and installed applications to create executable App
WorkEnvs with a runtime, UI observation, interaction interfaces, and reset
support.
Together, these resources supply the three Contexts used for WorkEnv
assembly and task synthesis in Phase~II.
Figure~\ref{fig:skill-distributions} summarizes skill-library coverage.
Appendix~\ref{app:context-construction-details} provides construction details
and complexity analysis; Appendix~\ref{sec:workspace-scale-quality} reports
resource coverage.

\subsubsection{Long-Horizon Task Synthesis}
\label{sec:task-generation}

Phase~II synthesizes task $T_i$ from the three Contexts in $C_i$
(Figure~\ref{fig:WorkRover-overview}). A construction profile sets the target
edit scope (files), interaction horizon (turns), skill complexity (number and
types of combined skills), distractor ratio, and cross-environment dependencies
(Table~\ref{tab:task-construction-profiles};
Appendix~\ref{app:task-construction-protocol}).
We profile selected file and app instances for task-relevant semantics,
relations, and executability. The Harness $H$ then assembles compatible
files, apps, and capabilities into a WorkEnv.
Starting from a job skill, the Skill Matcher fits the task scope to the
WorkEnv's available resources and combines job, file, and app skills into
an internal workflow whose stages depend on earlier outputs or state changes.
Its dependency graph guides preparation of the Starter WorkEnv $x_{v_i}$,
including all required resources and marking editable and protected state.
Task instructions specify goals, materials, deliverables, and constraints.
An independent Demonstrator Agent executes the task from $x_{v_i}$; success
confirms feasibility and yields a Reference State $x_i^\star$.
The Task Instruction, Starter WorkEnv, and Reference State form
the candidate, with state differences providing evidence of completion.
The workflow and reference execution must not add requirements or exclude
other valid solutions. We remove inputs to test dependency necessity;
independently confirmed compatibility problems trigger FCR
(Section~\ref{sec:workspace-coevolution}). Candidates passing construction
checks proceed to Phase~III for Verifier synthesis and validation.

\subsubsection{Verifier Synthesis and Validation}
\label{sec:harness-verification}

Phase~III synthesizes and validates the Verifier $V_i$
(Figure~\ref{fig:WorkRover-overview}). We derive an Evaluation Contract from
the task instructions and agent-visible workspace conditions, fixing its
criteria before inspecting the Reference State. The Contract specifies
expected outcomes, allowed edits, protected state, critical criteria, and
admissible evidence. We use it to generate a rubric and verification rules,
binding each criterion to evidence from files, application states, or state
differences.
\emph{Infrastructure Validity} returns $\mathit{invalid}$ when infrastructure
faults prevent evaluation; \emph{Canonical State Diff} checks state changes
against the Contract; \emph{Functional Checks} test artifact content and
consistency; and \emph{Semantic Judge} assesses criteria unresolved by
executable checks. Valid evaluations pass only if the weighted outcome score
meets its threshold and all mandatory checks pass; otherwise they fail.
We calibrate $V_i$ to reject the Starter State and negative cases that violate
the Contract, while accepting the Reference State and independently obtained
valid alternatives. Independently confirmed compatibility problems may
trigger FCR, followed by recalibration under the same fixed Contract.
We then freeze $V_i$ for a separate, one-time admission test; candidates that
fail are rejected without feedback for further repair.
Passing candidates become admitted packages $\mathcal{P}_i$ containing the
Task Instruction, Starter WorkEnv, Reference State, rubric, and evidence,
with accepted execution traces $\tau_i$ retained for training.
See Appendix~\ref{app:verifier-specification} for scoring and validation details.


\subsection{Recursive Environment Evolution}
\label{sec:workspace-coevolution}

The Work Context, Task, Harness, and Verifier must work together for a task
to run and be evaluated correctly. \emph{Factorial Compatibility Replay}
(FCR) repairs mismatches revealed during construction or later agent
rollouts. Since changing one component can affect the others, FCR tests
proposed changes both individually and jointly, then selects the smallest
validated repair that preserves the task's original intent.

\paragraph{Diagnosis and proposals.}
Independent diagnostics first distinguish environment problems from mistakes
in the agent's decisions. Agent mistakes alone do not trigger environment
changes. For an environment problem, the Harness records the affected
versions and dependencies, keeps task requirements and validation checks
fixed, and freezes one proposed repair per eligible component:
$\Delta C$, $\Delta T$, $\Delta H$, and $\Delta V$ as applicable.
The Evaluation Contract remains fixed once grounded in the task requirements.
Edits visible to the agent must preserve information boundaries, including
keeping hidden evaluation evidence hidden. These edits must also pass checks
by a fixed panel of solvers for unintended task simplification.

\paragraph{Harness screening and replay.}
For $d\leq4$ eligible components, each combination
$\chi\in\{0,1\}^{d}$ specifies which proposals to apply: $1$ applies a proposal
and $0$ omits it. This gives $2^d$ combinations, at most 16.
Before rebuilding packages or running rollouts, the Harness removes
combinations with confirmed conflicts in dependencies, resource bindings, or
requirements; uncertain cases proceed to validation. Each remaining combination is expanded
into a dependency-closed update set
$\Gamma(\chi)=\operatorname{Close}(\chi)$. This set includes the selected
repairs and all additional updates needed to keep dependent components
compatible, even if those components were not directly selected for repair.
The rebuilt packages are then replayed with the same agent policies,
execution budgets, and random seeds. Comparing individual and joint repairs reveals
whether some changes work only when applied together. If only the Verifier
changes, all alternatives evaluate the same recorded evidence.

\paragraph{Repair selection.}
Independent checks must confirm that the original problem is resolved,
required outcomes and information boundaries are preserved, and the repair
introduces no new problems in affected task packages. These checks do not use agent success or scores
from the proposed Verifier as evidence of resolution. Once a Verifier exists,
the repair must also pass the fixed Contract checks and tests confirming that
the Verifier rejects known failures and accepts valid alternative solutions.
Among the tested repair combinations that pass these checks, FCR selects
$\chi^\star$ whose full update set $\Gamma(\chi^\star)$ changes the fewest
component versions, counting updates required by dependencies. Ties favor
fewer rebuilt dependencies, then lower cost.

\paragraph{Confirmation and update.}
Fresh rollouts with new random seeds confirm the selected repair. A candidate
that has not yet been admitted to the training collection repeats validation
from the earliest affected construction step and must pass initial admission.
For an already admitted package, all validated changes take effect together
after confirmation. If repair selection or confirmation fails, FCR discards an unadmitted
candidate or keeps the previous admitted version. Later rollouts can reveal
new mismatches and initiate another repair cycle.
See Appendices~\ref{app:fcr-mechanics} and~\ref{app:fcr-evaluation}
for details.

\begin{figure}[!t]
    \centering
    \includegraphics[width=\linewidth]{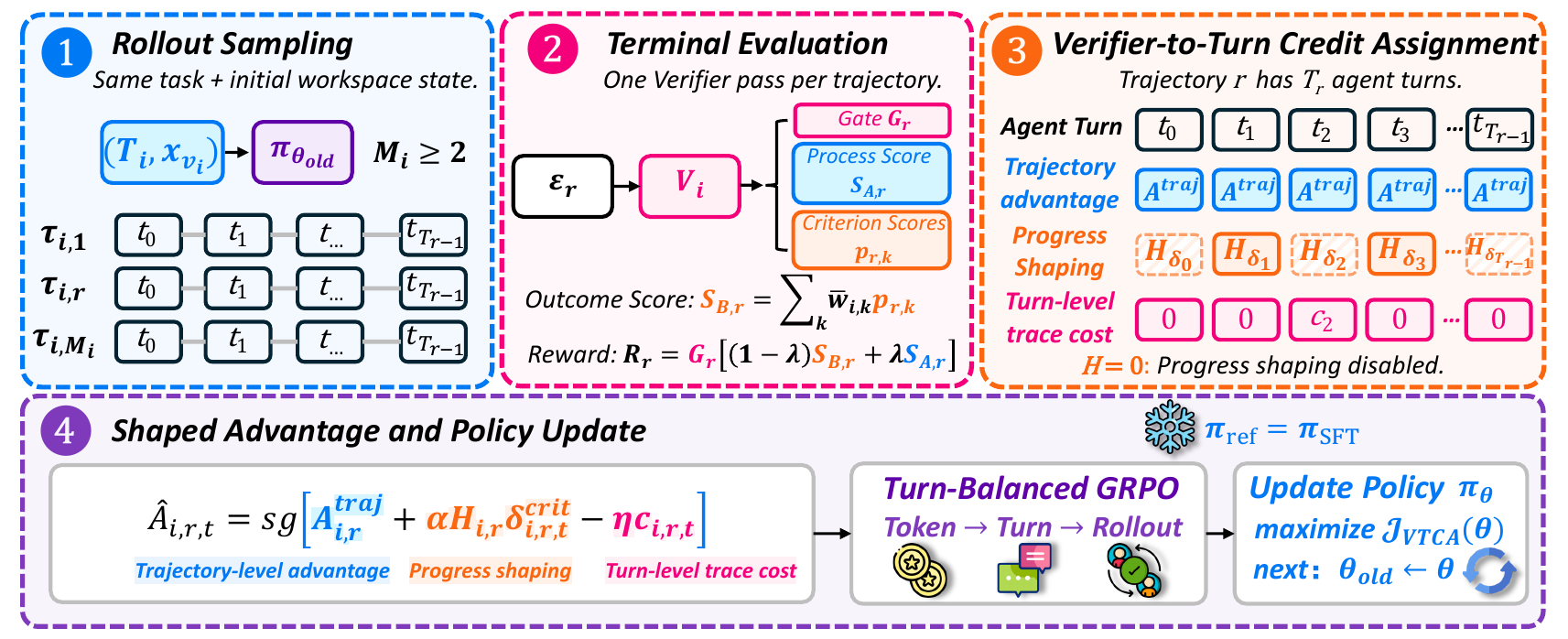}
    \captionsetup{justification=justified,singlelinecheck=false}
    \setlength{\abovecaptionskip}{4pt}
    \setlength{\belowcaptionskip}{0pt}
    \caption{\justifying\textbf{Overview of VTCA.} Same-task rollouts yield terminal
    rewards and checked progress. Bounded prerequisite allocations
    and local trace costs shape turn-balanced GRPO advantages.}
    \label{fig:vtca-training}
\end{figure}

\subsection{Verifier-to-Turn Credit Assignment}
\label{sec:training}

Terminal rewards score an entire rollout but provide little guidance about
which turns made progress. VTCA adds feedback from verified workspace changes
and recorded dependencies
(Figure~\ref{fig:vtca-training}).

\paragraph{Trajectory supervision.}
We initialize the policy with verified demonstrations using SFT, then sample
RL rollouts for the same task and initial state. Among rollouts with valid
evaluations, terminal rewards combine Verifier outcome scores and trace scores.
Rewards are zero if any critical criterion or integrity, protected-state, or
safety check fails. We standardize rewards within each group to obtain the
trajectory advantage $A^{\mathrm{traj}}_{i,r}$ for task $i$ and rollout $r$.

\paragraph{Measuring progress.}
Frozen deterministic checks~\citep{tao2026trace} score supported criteria before
and after each turn. We subtract the earlier score from the later one, so
increases indicate progress and decreases regression. If either score is
unavailable, we emit no local signal. Missing observations are never treated
as zero completion. Checks run on recorded states without intermediate Judge
calls.

\paragraph{Assigning turn credit.}
At least half of each positive increment stays with the turn that produced it.
The rest is divided equally among earlier turns whose tool outputs or exact
artifact versions were directly used to produce that progress, as confirmed
by recorded dependencies. For example, a turn that reads a table may share
credit when a later turn uses that returned table to compute a checked result.
Merely opening a file does not qualify. If no earlier turn qualifies, all
credit stays local. Negative increments remain entirely with the turn that
produced them.

\paragraph{Policy update.}
We combine these allocations using normalized criterion weights $\bar w_{i,k}$
to obtain criterion-based credit $\delta_{i,r,t}^{\mathrm{crit}}$ for turn $t$.
We add this credit to the trajectory advantage and subtract a local trace cost
$c_{i,r,t}$ for tool errors or unproductive repetition:
{\small
\begin{equation}
    \widehat A_{i,r,t}=\operatorname{sg}\!\left[
    A_{i,r}^{\mathrm{traj}}
    +\alpha H_{i,r}\delta_{i,r,t}^{\mathrm{crit}}
    -\eta c_{i,r,t}\right],
    \qquad \alpha,\eta\geq0,
    \label{eq:vtca-turn-advantage}
\end{equation}
}
Here $\alpha$ and $\eta$ weight progress credit and local trace cost, and
$\operatorname{sg}$ stops gradients. The gate $H_{i,r}$ is 1 only when
integrity, protected-state, and safety checks pass. An unfinished rollout can
retain progress credit, while violations disable it for every turn.
Clipped GRPO~\citep{shao2024deepseekmath} averages token contributions within
each turn, then turns within each trajectory, so lengthy turns do not dominate
the update.
Setting $\alpha=\eta=0$ recovers our episode-only baseline.
The reliability of this shaping signal depends on checker coverage and
dependency records (Appendix~\ref{app:vtca-details}).

\begingroup
\definecolor{MainTableGroupFill}{RGB}{235,235,235}
\definecolor{MainTableStripeFill}{RGB}{246,246,246}
\setlength{\textfloatsep}{8pt plus 2pt minus 2pt}
\setlength{\intextsep}{6pt plus 2pt minus 2pt}

\section{Experiments}
\label{sec:experiments}


\subsection{Experimental Setup}
\label{sec:experimental-setup}

\paragraph{Model and training data.}
We post-train Qwen3.5-35B-A3B~\citep{qwen2026qwen35} to obtain
\textsc{WorkRover}-35B-A3B. Supervised fine-tuning (SFT) uses 30,115
trajectories generated by Qwen3.5-397B-A17B in synthesized workspaces;
all pass task verification and quality checks. For reinforcement learning (RL),
we independently synthesize 20,000 task--Verifier pairs and screen each pair
with four Qwen3.5-35B-A3B pilot rollouts. Consistency filtering retains
3,200 pairs for training. Starting from the SFT checkpoint, the model collects
online rollouts on these tasks and learns with VTCA-GRPO
(Sec.~\ref{sec:training}).
Subset-selection details are in
Appendix~\ref{app:experimental-setup}.

\paragraph{Evaluation.}
We evaluate transfer to five external benchmarks: JobBench~\citep{li2026jobbenchaligningagent}
for occupational tasks, GDPval~\citep{artificialanalysis2026gdpvalaa} for
professional deliverables, AutomationBench~\citep{shepard2026automationbench}
for cross-application workflows, WorkBuddy Bench
(Office)~\citep{tencent2026workbuddybench} for office workflows, and
OfficeQA Pro~\citep{opsahlong2026officeqapro} for document-grounded question
answering. We use each benchmark's official evaluator and aggregate metric,
reporting GDPval in Elo and the other four scores as percentages.
Table~\ref{tab:main-results} compares \textsc{WorkRover} with its backbone,
other 27--35B open-weight models, open-weight models above 400B, and
proprietary models. Published reference scores retain their original task
subsets, evaluators, and harnesses. Our training and evaluation setup is
described in Appendix~\ref{app:experimental-setup}.

\subsection{Main Results}
\label{sec:agent-performance}

\textsc{WorkRover}-35B-A3B improves on its Qwen3.5-35B-A3B backbone across
all five external benchmarks (Table~\ref{tab:main-results}). Among the
percentage-based scores, the largest gains are on OfficeQA Pro (+29.32
percentage points) and JobBench (+26.27 points), followed by AutomationBench
(+22.67 points) and WorkBuddy Bench (Office) (+13.63 points). GDPval v2
increases from 804.75 to 1218.00 Elo, a gain of 413.25. These improvements
span document reasoning, occupational tasks, professional deliverables, and
application workflows, supporting transfer beyond the synthesized training
workspaces.

Among the listed 27--35B open-weight models, \textsc{WorkRover} has the
highest reported score on every benchmark. Occamy-1.0 has the
second-highest reported score in this group on each benchmark.
The largest percentage-point gap between the two models is on
WorkBuddy Bench (Office): 74.13\% versus 66.65\%, a margin of 7.48 points.
The JobBench margin is 2.62 points (36.89\% versus 34.27\%), while the
leads on AutomationBench and OfficeQA Pro are smaller, at 1.07 and 0.77
points, respectively.

Compared with larger and proprietary models, \textsc{WorkRover}'s relative
performance varies across benchmarks. It exceeds GLM-5.2 and MiniMax-M3 on
both AutomationBench and OfficeQA Pro, and Kimi K3 on WorkBuddy Bench
(Office) (74.13\% versus 72.18\%). Gaps remain on occupational work and
deliverable quality: its JobBench score of 36.89\% is below GPT-5.5
(38.30\%) and Kimi K3 (54.27\%), and its GDPval score of 1218.00 Elo is
below all listed open-weight models above 400B (1303.96--1572.42 Elo).
WorkBuddy performance also remains below GPT-5.5 (86.05\%) and Claude
Opus 4.7 (92.51\%). Comparisons with reference models use their reported
scores under source-specific evaluation settings.

\begin{table}[!t]
\centering
\small
\definecolor{MainTableGain}{RGB}{255,69,0}
\definecolor{MainTableDrop}{RGB}{0,149,173}
\setlength{\tabcolsep}{3pt}
\renewcommand{\arraystretch}{1.15}
\newcommand{\TableModelLogo}[2][]{%
  \makebox[1.1em][c]{\raisebox{-0.17em}[0pt][0pt]{%
    \includegraphics[width=0.98em,height=0.98em,keepaspectratio,#1]{#2}}}%
  \hspace{0.25em}%
}
\newcommand{\MainScoreCell}[2]{%
  \makebox[2.55em][r]{$#1$}%
  \makebox[2.75em][l]{$\vphantom{#1}_{#2}$}%
}
\newcommand{\MainEloCell}[2]{%
  \makebox[3.55em][r]{$#1$}%
  \makebox[3.15em][l]{$\vphantom{#1}_{#2}$}%
}
\newcommand{\MainScoreUp}[1]{\textcolor{MainTableGain}{\uparrow #1}}
\newcommand{\MainScoreDown}[1]{\textcolor{MainTableDrop}{\downarrow #1}}
\newcommand{\MainTableHeader}[1]{%
  \begin{tabular}[c]{@{}c@{}}#1\end{tabular}%
}
\newcommand{\MainTableSectionRule}{%
  \noalign{\vskip 0.5pt}%
  \arrayrulecolor{black}\hline%
  \noalign{\vskip 1pt}%
}
\caption{\justifying Benchmark results.
\textbf{Bold} and \underline{underlined} mark the best and second-best
scores per model group. Subscripts show changes from
Qwen3.5-35B-A3B: \textcolor{MainTableGain}{$\uparrow$ increases} and
\textcolor{MainTableDrop}{$\downarrow$ decreases} (Elo or percentage points).
Higher scores are better.
Details in Appendix~\ref{app:experimental-setup}.}
\label{tab:main-results}
\TableOuterRules{%
\resizebox{\linewidth}{!}{%
\begin{tabular}{lccccc}
\noalign{\vskip 2pt}
\textbf{Model} &
\MainTableHeader{\textbf{JobBench}\\(\%) $\uparrow$} &
\MainTableHeader{\textbf{GDPval}\\(Elo) $\uparrow$} &
\MainTableHeader{\textbf{Automation}\\\textbf{Bench} (\%) $\uparrow$} &
\MainTableHeader{\textbf{WorkBuddy}\\\textbf{Bench (Office)} (\%) $\uparrow$} &
\MainTableHeader{\textbf{OfficeQA}\\\textbf{Pro} (\%) $\uparrow$} \\
\MainTableSectionRule
\rowcolor{MainTableGroupFill}
\multicolumn{6}{c}{\textbf{\textit{Proprietary Models}}} \\
\TableModelLogo{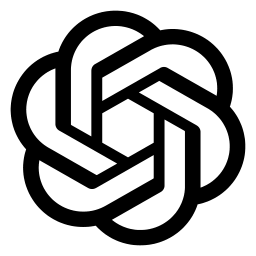} GPT-5.5 & \MainScoreCell{\underline{38.30}}{\MainScoreUp{27.68}} & \MainEloCell{\mathbf{1395.85}}{\MainScoreUp{591.10}} & \MainScoreCell{\underline{22.70}}{\MainScoreUp{16.70}} & \MainScoreCell{\underline{86.05}}{\MainScoreUp{25.55}} & \MainScoreCell{\mathbf{57.10}}{\MainScoreUp{37.55}} \\ 
\rowcolor{MainTableStripeFill}
\TableModelLogo{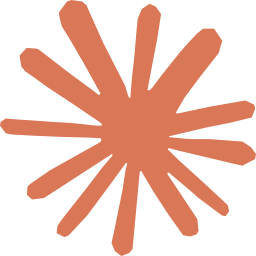} Claude Opus 4.7 & \MainScoreCell{\mathbf{44.50}}{\MainScoreUp{33.88}} & \MainEloCell{\underline{1395.78}}{\MainScoreUp{591.03}} & \MainScoreCell{9.90}{\MainScoreUp{3.90}} & \MainScoreCell{\mathbf{92.51}}{\MainScoreUp{32.01}} & \MainScoreCell{\underline{46.60}}{\MainScoreUp{27.05}} \\ 
\TableModelLogo{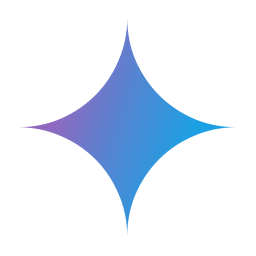} Gemini 3.1 Pro Preview & \MainScoreCell{15.90}{\MainScoreUp{5.28}} & \MainEloCell{903.94}{\MainScoreUp{99.19}} & \MainScoreCell{\mathbf{37.54}}{\MainScoreUp{31.54}} & \MainScoreCell{\text{--}}{} & \MainScoreCell{39.10}{\MainScoreUp{19.55}} \\ 
\MainTableSectionRule
\rowcolor{MainTableGroupFill}
\multicolumn{6}{c}{\textbf{\textit{Open-Weight Models (\textgreater{}400B total parameters)}}} \\
\TableModelLogo{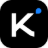} Kimi K3 & \MainScoreCell{\mathbf{54.27}}{\MainScoreUp{43.65}} & \MainEloCell{\mathbf{1572.42}}{\MainScoreUp{767.67}} & \MainScoreCell{\mathbf{46.67}}{\MainScoreUp{40.67}} & \MainScoreCell{72.18}{\MainScoreUp{11.68}} & \MainScoreCell{\underline{63.30}}{\MainScoreUp{43.75}} \\ 
\rowcolor{MainTableStripeFill}
\TableModelLogo{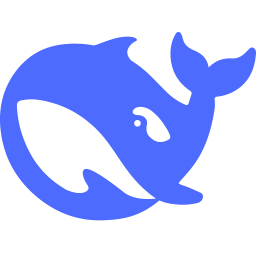} DeepSeek-V4-Pro & \MainScoreCell{\underline{54.10}}{\MainScoreUp{43.48}} & \MainEloCell{\underline{1493.31}}{\MainScoreUp{688.56}} & \MainScoreCell{\underline{31.80}}{\MainScoreUp{25.80}} & \MainScoreCell{\underline{78.71}}{\MainScoreUp{18.21}} & \MainScoreCell{\mathbf{65.40}}{\MainScoreUp{45.85}} \\ 
\TableModelLogo{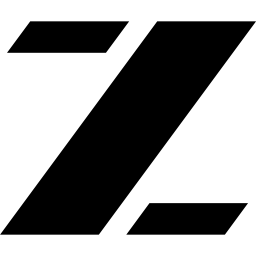} GLM-5.2 & \MainScoreCell{43.40}{\MainScoreUp{32.78}} & \MainEloCell{1417.62}{\MainScoreUp{612.87}} & \MainScoreCell{26.17}{\MainScoreUp{20.17}} & \MainScoreCell{\mathbf{79.57}}{\MainScoreUp{19.07}} & \MainScoreCell{41.40}{\MainScoreUp{21.85}} \\ 
\rowcolor{MainTableStripeFill}
\TableModelLogo{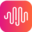} MiniMax-M3 & \MainScoreCell{\text{--}}{} & \MainEloCell{1303.96}{\MainScoreUp{499.21}} & \MainScoreCell{15.90}{\MainScoreUp{9.90}} & \MainScoreCell{76.30}{\MainScoreUp{15.80}} & \MainScoreCell{45.10}{\MainScoreUp{25.55}} \\ 
\MainTableSectionRule
\rowcolor{MainTableGroupFill}
\multicolumn{6}{c}{\textbf{\textit{Open-Weight Models (27--35B total parameters)}}} \\
\TableModelLogo{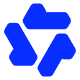} Qwen3.6-27B & \MainScoreCell{21.80}{\MainScoreUp{11.18}} & \MainEloCell{1068.87}{\MainScoreUp{264.12}} & \MainScoreCell{9.67}{\MainScoreUp{3.67}} & \MainScoreCell{65.65}{\MainScoreUp{5.15}} & \MainScoreCell{44.36}{\MainScoreUp{24.81}} \\ 
\rowcolor{MainTableStripeFill}
\TableModelLogo{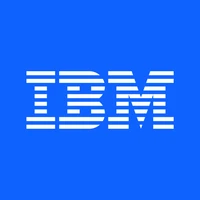} Granite-4.2-30B & \MainScoreCell{1.64}{\MainScoreDown{8.98}} & \MainEloCell{721.01}{\MainScoreDown{83.74}} & \MainScoreCell{0.02}{\MainScoreDown{5.98}} & \MainScoreCell{54.18}{\MainScoreDown{6.32}} & \MainScoreCell{12.03}{\MainScoreDown{7.52}} \\ 
\TableModelLogo{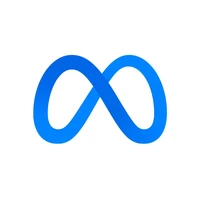} Muse-Glimmer-30B & \MainScoreCell{11.82}{\MainScoreUp{1.20}} & \MainEloCell{892.72}{\MainScoreUp{87.97}} & \MainScoreCell{6.83}{\MainScoreUp{0.83}} & \MainScoreCell{57.45}{\MainScoreDown{3.05}} & \MainScoreCell{24.06}{\MainScoreUp{4.51}} \\ 
\rowcolor{MainTableStripeFill}
\TableModelLogo{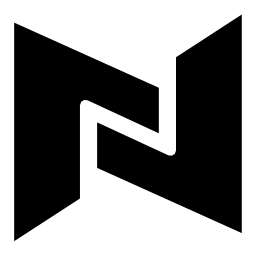} Nex-N2-mini & \MainScoreCell{3.82}{\MainScoreDown{6.80}} & \MainEloCell{1097.03}{\MainScoreUp{292.28}} & \MainScoreCell{5.70}{\MainScoreDown{0.30}} & \MainScoreCell{25.87}{\MainScoreDown{34.63}} & \MainScoreCell{26.32}{\MainScoreUp{6.77}} \\ 
\TableModelLogo{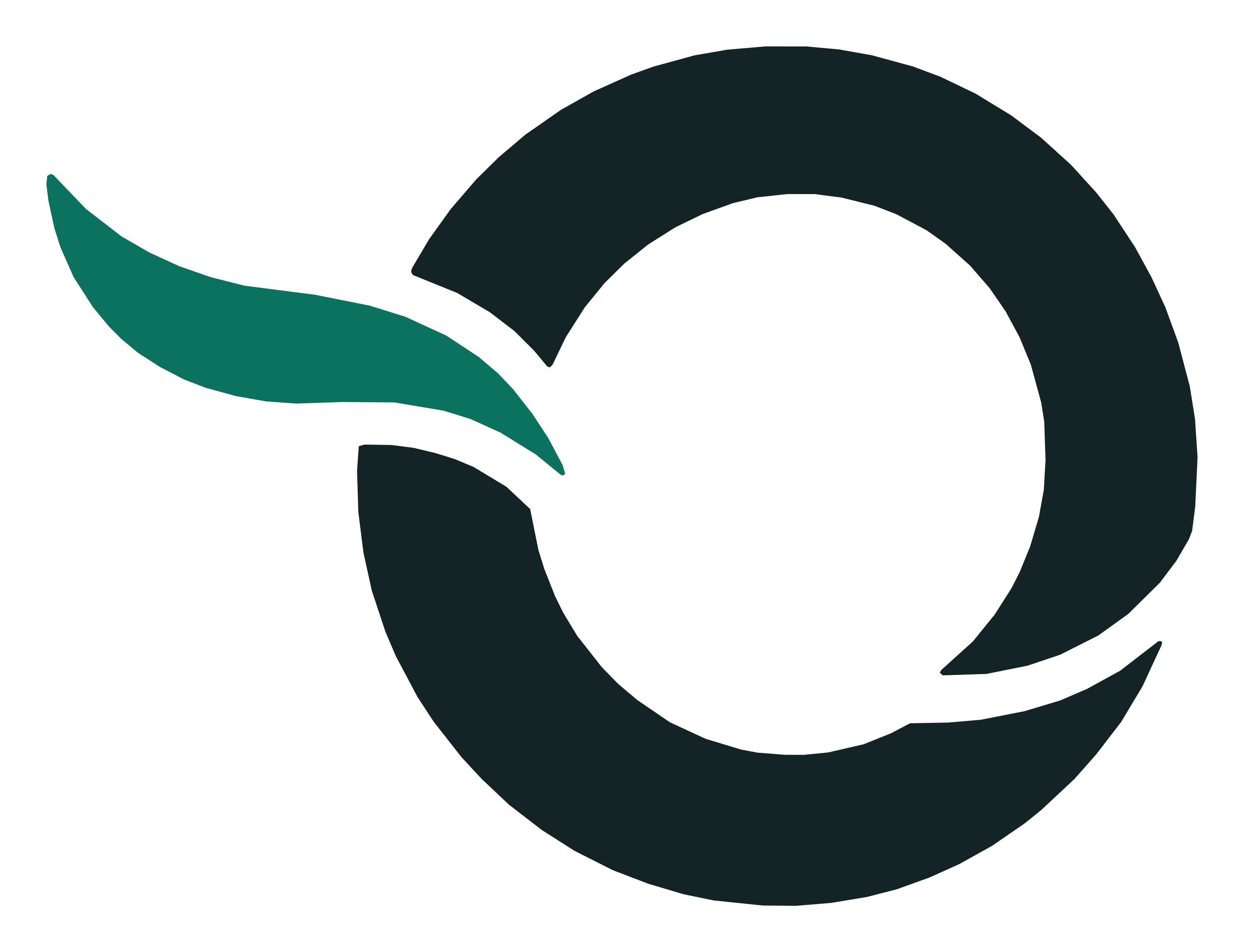} Occamy-1.0 & \MainScoreCell{\underline{34.27}}{\MainScoreUp{23.65}} & \MainEloCell{\underline{1128.00}}{\MainScoreUp{323.25}} & \MainScoreCell{\underline{27.60}}{\MainScoreUp{21.60}} & \MainScoreCell{\underline{66.65}}{\MainScoreUp{6.15}} & \MainScoreCell{\underline{48.10}}{\MainScoreUp{28.55}} \\ 
\rowcolor{MainTableStripeFill}
\TableModelLogo{figures/logos/model-qwen.png} Qwen3.5-35B-A3B & \MainScoreCell{10.62}{} & \MainEloCell{804.75}{} & \MainScoreCell{6.00}{} & \MainScoreCell{60.50}{} & \MainScoreCell{19.55}{} \\ 
\MainTableSectionRule
\TableModelLogo[viewport=200 180 732 532,clip]{figures/logos/workrover/workrover-logo-transparent.png} \textbf{\textsc{WorkRover}-35B-A3B (Ours)} & \MainScoreCell{\mathbf{36.89}}{\MainScoreUp{26.27}} & \MainEloCell{\mathbf{1218.00}}{\MainScoreUp{413.25}} & \MainScoreCell{\mathbf{28.67}}{\MainScoreUp{22.67}} & \MainScoreCell{\mathbf{74.13}}{\MainScoreUp{13.63}} & \MainScoreCell{\mathbf{48.87}}{\MainScoreUp{29.32}} \\ 
\noalign{\vskip 2pt}
\end{tabular}%
}
}
\end{table}

\subsection{Ablations and Analysis}
\label{sec:ablation-studies}

\subsubsection{Verifier-to-Turn Credit Assignment}
\label{sec:training-results}
With the SFT checkpoint, task and seed schedules, and rollout budget fixed,
we compare VTCA with episode-only GRPO (E-GRPO;
Figure~\ref{fig:vtca-main-results}). Both RL methods improve on SFT across
all five benchmarks. VTCA further outperforms E-GRPO on every benchmark.
The largest percentage-point gains are on AutomationBench (+5.7) and
OfficeQA Pro (+5.4), while GDPval v2 improves by 43 Elo.
These results support giving individual turns
feedback beyond the final trajectory outcome. Protocol and component
controls are in
Appendices~\ref{app:training-results} and~\ref{app:vtca-details}.


\begin{figure}[!htbp]
\centering
\includegraphics[width=\linewidth]{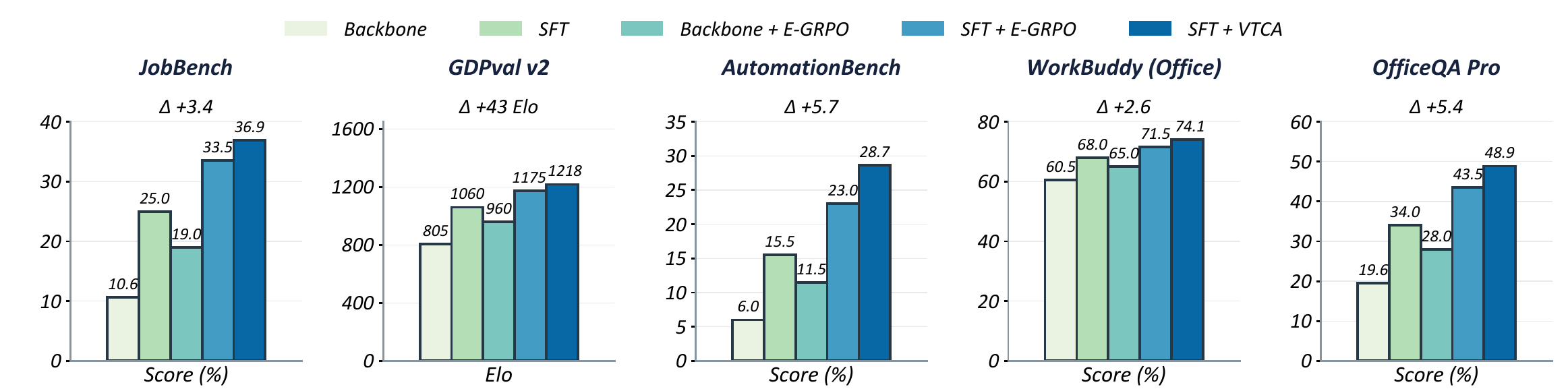}
\caption{\RaggedRight 35B-A3B comparison under matched rollout budgets.
Backbone denotes Qwen3.5-35B-A3B before our SFT or RL.
Gains compare SFT + VTCA with SFT + E-GRPO.
GDPval v2 uses Elo; other scores are percentages.
}
\label{fig:vtca-main-results}
\end{figure}

\subsubsection{Recursive Environment Evolution}
\label{sec:coevolution-analysis}

We compare repair methods using Qwen3.5-35B-A3B under matched proposal and
replay budget caps. Safe Resolution Rate (SRR) measures how often a defect
is fixed without causing new failures in related tasks; repairs must also
pass fresh-seed confirmation and a held-out audit. FCR raises SRR on
naturally occurring defects from 60.7\% with full regeneration to 70.3\%
(Figure~\ref{fig:fcr-result-summary}(b)). It also reduces changes to task
intent from 14.3\% to 1.3\% while using fewer model tokens (relative cost:
1.00 versus 1.34).

\paragraph{Choosing compatible edits.}
With repair proposals fixed, Table~\ref{tab:fcr-shared-proposal-main} tests
selection separately from proposal generation. Compared with applying all
edits (Apply-all), FCR raises SRR by 9.0 percentage points on natural defects
and 10.4 points on injected defects, and reduces cross-task regression from
7.0\% to 0.7\%. Testing edit combinations thus improves repair reliability,
at a higher model-token cost (1.00 versus 0.59).

\paragraph{Joint repair and repeated evolution.}
With proposals fixed, 48--74\% of safely resolved cases require joint edits,
depending on the component pair: no smaller subset of the selected edits
passes the same repair checks (Figure~\ref{fig:fcr-result-summary}(a)).
Across three generations, SRR stays at 70.3--72.4\% and audit pass rate at
93.8--94.6\% (panel~c), showing little change in repair quality across these
rounds. FCR's task success rate also increases over five repair iterations (panel~d).

\begin{figure}[!t]
\centering
\includegraphics[width=\linewidth]{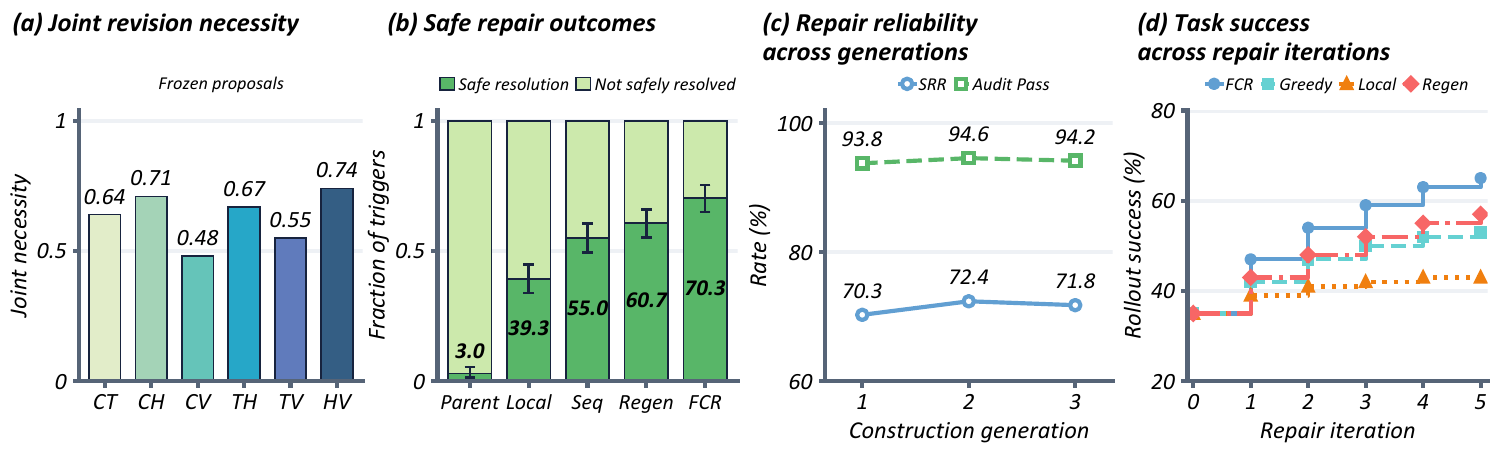}
\caption{\textbf{FCR mechanism, stability, and evolution.}
(a) Pairwise joint necessity (C: Context; T: Task; H: Harness; V: Verifier).
(b) Natural-trigger outcomes. (c) Repair reliability across three generations,
measured by SRR and sealed-audit pass rate.
(d) Task success across five repair iterations.}
\label{fig:fcr-result-summary}
\end{figure}

\paragraph{Repairing initial candidates.}
FCR also improves initial construction. With the candidate set and repair
budget caps fixed, synthesis pass rate reaches 74.8\%, compared with 49.8\%
before repair and 68.8\% with full regeneration. Repair therefore recovers
more tasks that pass validation from the same initial pool.

\subsubsection{Construction Pipeline}
\label{sec:construction-ablation-results}
We remove one construction component at a time
(Table~\ref{tab:construction-main-ablation}). Task yield measures the fraction
of initial candidates that pass validation. The largest drops come from removing relation-aware expansion
(74.8\% to 52.6\%) or validation-guided rewriting (74.8\% to 49.8\%). These
results highlight the value of following file dependencies and repairing
construction errors.

Reference execution and agent success decline as well. Removing relation-aware
expansion lowers them from 95.1\% and 58.7\% to 77.9\% and 41.8\%,
respectively. Removing file--skill grounding also reduces both measures,
supporting the need to match each file with operations that can use it.
The semantic-only baseline performs worse on all three
measures, suggesting that topical similarity alone is insufficient for
assembling executable workspaces. Success rates use the same frozen agent
but are measured on each variant's retained workspaces, so they can also
reflect differences in which tasks are admitted.

\begin{table}[!htbp]
\centering
\newcommand{\PairedTableHeaderStrut}{%
  \vphantom{\shortstack{\textbf{Task}\\\textbf{reg.}$\uparrow\downarrow$}}%
}
\newcommand{\FCRBodyStrut}{\rule[-4.67pt]{0pt}{12.7pt}}
\begin{minipage}[t]{0.48\linewidth}
\vspace{0pt}
\setlength{\abovecaptionskip}{0pt}
\caption{\RaggedRight Maintenance with shared frozen proposals.
Matched budget caps; rates (\%); model-token cost relative to FCR.}
\label{tab:fcr-shared-proposal-main}
\end{minipage}\hfill
\begin{minipage}[t]{0.50\linewidth}
\vspace{0pt}
\setlength{\abovecaptionskip}{0pt}
\caption{\RaggedRight Construction ablations (\%). Mean $\pm$ workspace-clustered
95\% interval half-width.}
\label{tab:construction-main-ablation}
\end{minipage}
\par\vspace{2pt}
\begin{minipage}[t]{0.48\linewidth}
\vspace{0pt}
\centering
\scriptsize
\setlength{\tabcolsep}{1pt}
\renewcommand{\arraystretch}{1}
\TableOuterRules{%
\begin{tabular*}{\linewidth}{@{\extracolsep{\fill}}lcccc@{}}
\PairedTableHeaderStrut\textbf{Method} & \shortstack{\textbf{Natural}\\\textbf{SRR}$\uparrow$} &
\shortstack{\textbf{Controlled}\\\textbf{SRR}$\uparrow$} &
\shortstack{\textbf{Task}\\\textbf{reg.}$\downarrow$} &
\shortstack{\textbf{Rel.}\\\textbf{cost}$\downarrow$} \\
\hline
\FCRBodyStrut Local-only & 42.0 & 45.4 & 3.2 & 0.43 \\
\FCRBodyStrut Greedy-sequential & 58.0 & 63.1 & 2.7 & 0.78 \\
\FCRBodyStrut Apply-all & 64.0 & 70.8 & 7.0 & 0.59 \\
\FCRBodyStrut\textbf{FCR} & \textbf{73.0} & \textbf{81.2} & \textbf{0.7} & 1.00 \\
\end{tabular*}%
}
\end{minipage}\hfill
\begin{minipage}[t]{0.50\linewidth}
\vspace{0pt}
\centering
\scriptsize
\setlength{\tabcolsep}{1pt}
\renewcommand{\arraystretch}{1.05}
\TableOuterRules{%
\begin{tabular*}{\linewidth}{@{\extracolsep{\fill}}lccc@{}}
\PairedTableHeaderStrut\textbf{Construction} & \shortstack{\textbf{Task}\\\textbf{yield}} &
\shortstack{\textbf{Reference}\\\textbf{exec.}} & \shortstack{\textbf{Success}\\\textbf{Rate}} \\
\hline
\textbf{\textsc{WorkRover} (full)} & $74.8_{\pm2.20}$ & $95.1_{\pm1.10}$ & $58.7_{\pm2.60}$ \\
\textit{w/o Job--Skill} & $67.2_{\pm2.38}$ & $91.7_{\pm1.48}$ & $53.1_{\pm2.78}$ \\
\textit{w/o File--Skill} & $60.9_{\pm2.47}$ & $84.2_{\pm2.06}$ & $47.4_{\pm2.92}$ \\
\textit{w/o Relation Expansion} & $52.6_{\pm2.53}$ & $77.9_{\pm2.52}$ & $41.8_{\pm3.11}$ \\
\textit{w/o Validation Rewrite} & $49.8_{\pm2.53}$ & $80.6_{\pm2.47}$ & $44.5_{\pm3.22}$ \\
\hline
Semantic-only & $48.7_{\pm2.53}$ & $73.6_{\pm2.78}$ & $38.6_{\pm3.19}$ \\
\end{tabular*}%
}
\end{minipage}
\end{table}

\paragraph{Rewriting on the same initial tasks.}
To control for differences in task selection, we compare filtering with and
without rewriting on the same 10,000 initial task intents. Both variants preserve
the required outcomes and use the same frozen agent. Adding rewriting raises
admission from 49.8\% to 74.8\% and end-to-end success from 22.2\% to 43.9\%.
Reference execution also rises from 80.0\% to 92.0\%, and Verifier validity
from 70.0\% to 87.0\%. Unconstructed, rejected, and budget-exhausted intents
remain in the denominator and contribute no successful completion. The
resulting gains therefore measure the full pipeline, from initial construction
through agent execution, on the same starting task set.
\endgroup


\section{Conclusion}
\label{sec:discussion}
\label{sec:conclusion}

\textsc{WorkRover} uses execution feedback to improve both long-horizon
cowork agents and the environments used to train them. The Harness keeps
work contexts, tasks, and verifiers aligned during construction and repair.
FCR revises inconsistent components while preserving task requirements,
and VTCA uses verified workspace progress to guide learning at individual
turns. Experiments show that FCR increases the share of tasks that pass
validation and repairs defects with fewer unintended changes to task intent
and fewer regressions on related tasks. Under matched rollout budgets,
VTCA outperforms episode-only GRPO on all five benchmarks. These findings
support recursive environment evolution as a practical foundation for
reliable long-horizon cowork agent training.

\FloatBarrier
\begingroup
\setlength{\parskip}{0pt}
\setlength{\bibsep}{0pt plus 0.5pt}
\bibliographystyle{abbrvnat}
\bibliography{fsworld_references}
\endgroup

\appendix

\clearpage
\setlength{\parskip}{1.5pt plus 1pt minus 0.5pt}
\setlength{\textfloatsep}{10pt plus 2pt minus 2pt}
\setlength{\floatsep}{6pt plus 2pt minus 2pt}
\setlength{\intextsep}{6pt plus 2pt minus 2pt}
\setlength{\abovecaptionskip}{4pt}
\setlength{\belowcaptionskip}{0pt}
\setlength{\abovedisplayskip}{3.5pt plus 2pt minus 1pt}
\setlength{\belowdisplayskip}{3.5pt plus 2pt minus 1pt}
\setlength{\abovedisplayshortskip}{0pt plus 2pt}
\setlength{\belowdisplayshortskip}{2.5pt plus 2pt minus 1pt}
\renewcommand{\topfraction}{0.95}
\renewcommand{\bottomfraction}{0.9}
\renewcommand{\textfraction}{0.05}
\renewcommand{\floatpagefraction}{0.85}
\setcounter{topnumber}{4}
\setcounter{bottomnumber}{4}
\setcounter{totalnumber}{6}
\section{Extended Related Work}
\label{app:related-work}

\paragraph{Executable environments and task synthesis.}
Professional agent training requires executable tasks, recoverable state,
and observable outcomes across related artifacts. AppWorld provides
controllable application environments; EnvScaler supports programmatic
environment construction, while ClawGym materializes workspaces around task
and capability requirements
\citep{trivedi2024appworldcontrollableworldapps,
song2026envscalerscalingtoolinteractive,bai2026clawgymscalableframeworkbuilding}.
Agent-World uses capability diagnostics to expand tool/database environments
\citep{dong2026agentworld}. ToolForge constructs virtual-tool tasks, and
MetaTask and CLI-Universe emphasize executable validation
\citep{chen2025toolforgedatasynthesispipeline,chen2026toolforge,pan2026metataskturningterminaltask,
hua2026cliuniverse}. These approaches organize synthesis around task bundles,
environment expansion, or capability requirements. \textsc{WorkRover} grounds
task construction in file relations and professional skills, then assembles
Tasks and Verifiers within recoverable Work Contexts under a versioned Harness.
Shared contexts also make dependency tracking and revalidation necessary when
one component changes.

\paragraph{Grounded acceptance and verifier reliability.}
Acceptance checks must agree with the instruction and the evidence available
in the workspace. ANCHOR constructs instructions, environments, solutions,
and verifiers from a shared workflow specification \citep{ivanov2026anchor}.
FACET grounds artifacts in shared state and repairs inconsistencies while
preserving source intent \citep{shi2026facet}. Hacker--Fixer audits tests and
environments with adversarial cases while checking legitimate solutions
\citep{zhong2026hackerfixer}. Stronger rejection checks therefore need to be
considered alongside false rejection of valid alternatives. \textsc{WorkRover}
freezes instruction-grounded requirements and criterion--evidence bindings,
then evaluates implementations using violating mutants and independently
valid solutions. Reference artifacts supply recomputable anchors without
adding requirements. Infrastructure-invalid episodes are distinguished from
policy failures. Validation remains limited to the declared Contract and
the coverage of the controls.

\paragraph{Environment evolution and repair selection.}
Envs-FORGE studies joint revision of instructions, fixtures, solutions, tests,
and environments, while EchoVerse studies executable environment evolution
\citep{wu2026envsforge,pandya2026echoverse}. FACET routes failed validation
to implicated artifacts and repeats validation from a clean state
\citep{shi2026facet}. Program repair likewise studies patch minimization:
GenProg reduces test-passing patches through structural differencing and
delta debugging \citep{weimer2009genprog}. FCR freezes component-level
proposals over Context, Task, Harness, and Verifier before replaying their
combinations. Materialization includes dependency-induced changes, and promotion
checks affected retained tasks for regressions under fixed requirements.
The selected repair is the smallest admissible closed edit set among the
tested combinations; bounded enumeration does not establish a global minimum
over all possible repairs. This separates proposal generation, compatibility
selection, and validation of a shared environment revision.

\paragraph{Local progress in long-horizon learning.}
Methods differ in the evidence used to assign local feedback. GiGPO uses
state comparisons, while HiPER uses a planning hierarchy
\citep{feng2025gigpo,peng2026hiperhierarchicalreinforcementlearning}.
TRACE forms temporal differences in frozen-model reference-answer likelihoods,
with bounded look-ahead for delayed tool effects \citep{tao2026trace}.
TRCA combines deterministic transition-rubric scores with increments in
cumulative condition coverage \citep{zhang2026trca}. VTCA instead evaluates
successive observable workspace states with frozen checkers. Signed progress
can register both completion and setbacks in persistent, mutable artifacts.
Unsupported or unavailable evidence supplies no local signal, while terminal
evaluation remains applicable to supported task outcomes. These differences
concern the feedback source and its coverage; they do not by themselves
establish the comparative effectiveness of individual components.

\paragraph{Evidence attribution and shaping limits.}
STAMP maps verified citations to first-exposure actions; DRACO uses
judge-attributed steps with sign-preserving GRPO advantage redistribution
\citep{xu2026stamp,gandhi2026draco}. VTCA follows recorded direct use of exact
tool outputs or artifact versions, assigning at most half of positive progress
to eligible prerequisite turns and leaving negative increments local.
Access alone creates no dependency edge. Incorrect provenance can misallocate
credit, and missed links leave prerequisites unrepresented. Recorded use
therefore does not establish causal necessity. Potential-based reward shaping
preserves optimal policies under stated assumptions \citep{ng1999policyinvariance};
VTCA is formulated as workspace-grounded advantage shaping. Conservation of
checker increments before gating, costs, and clipping establishes neither
policy invariance nor an unbiased policy-gradient estimate.

\section{Data Construction and Case Studies}
\label{sec:appendix}

\subsection{Work Context Construction}
\label{app:context-construction-details}
For each source workspace $w$, the index $I(w)$ stores a typed graph
$G_w=(V_w,E_w)$ with files or versions as nodes and semantic, structural,
temporal, provenance, and workflow edges. Directory co-location is only a weak
structural signal. A job skill specifies its professional setting, workflow,
materials, and deliverables. The Skill Matcher retrieves related file regions,
then expands them through file and application skills with compatible inputs,
outputs, and interfaces. The Harness validates dependencies, permissions,
bindings, reset, recovery, and evidence capture. Recorded lineage and derived
task branches stay in one data split.

\paragraph{Construction complexity.}
Let $J$ be the number of job skills, $W$ the number of source workspaces,
$k$ the number of shortlisted regions per job skill, and $M\leq Jk$ the
number of candidates retained for environment assembly. Let
$C_{\mathrm{index}}$ and $C_{\mathrm{retrieve}}$ denote per-workspace indexing
and per-job-skill retrieval costs; $C_{\mathrm{agent}}$,
$C_{\mathrm{match}}$, and $C_{\mathrm{env}}$ denote per-candidate agent
inspection, skill matching, and environment-assembly costs. Excluding source
collection and one-time pool construction, the total cost of indexed
Work Context construction is
{\small
\begin{equation}
    O(WC_{\mathrm{index}})+O(JC_{\mathrm{retrieve}})
    +O\!\left(Jk(C_{\mathrm{agent}}+C_{\mathrm{match}})\right)
    +O(MC_{\mathrm{env}}).
    \label{eq:context-construction-cost}
\end{equation}
}
Agent inspection covers the $Jk$ shortlisted candidates, with cost
$O(JkC_{\mathrm{agent}})$, whereas exhaustive job--workspace inspection
incurs $O(JWC_{\mathrm{agent}})$. This comparison concerns agent inspection;
indexing, retrieval, skill matching, and environment assembly remain separate
costs. The index is reusable across job-skill queries, so its construction
cost is paid once for a fixed source collection.

\subsection{Synthesis Resources and Coverage}
\label{sec:workspace-scale-quality}
The inventory contains 21,444 app skills, 5,918 file skills, and 36,064 job
skills covering 317 roles. App and job skills receive one category label across
15 and 10 categories; file skills have 11,273 memberships across 22 categories.
Figure~\ref{fig:skill-distributions} shows the six largest categories per
library and groups the remainder as Other. Construction profiles describe
artifact diversity, file relations, workflow horizon, cross-root dependencies,
and noise; empirical difficulty is calibrated separately.

\subsection{Workspace Construction from Deep Research Sessions}
\label{app:deep-research-construction}
\begin{figure}[!htbp]
\centering
\includegraphics[width=\linewidth]{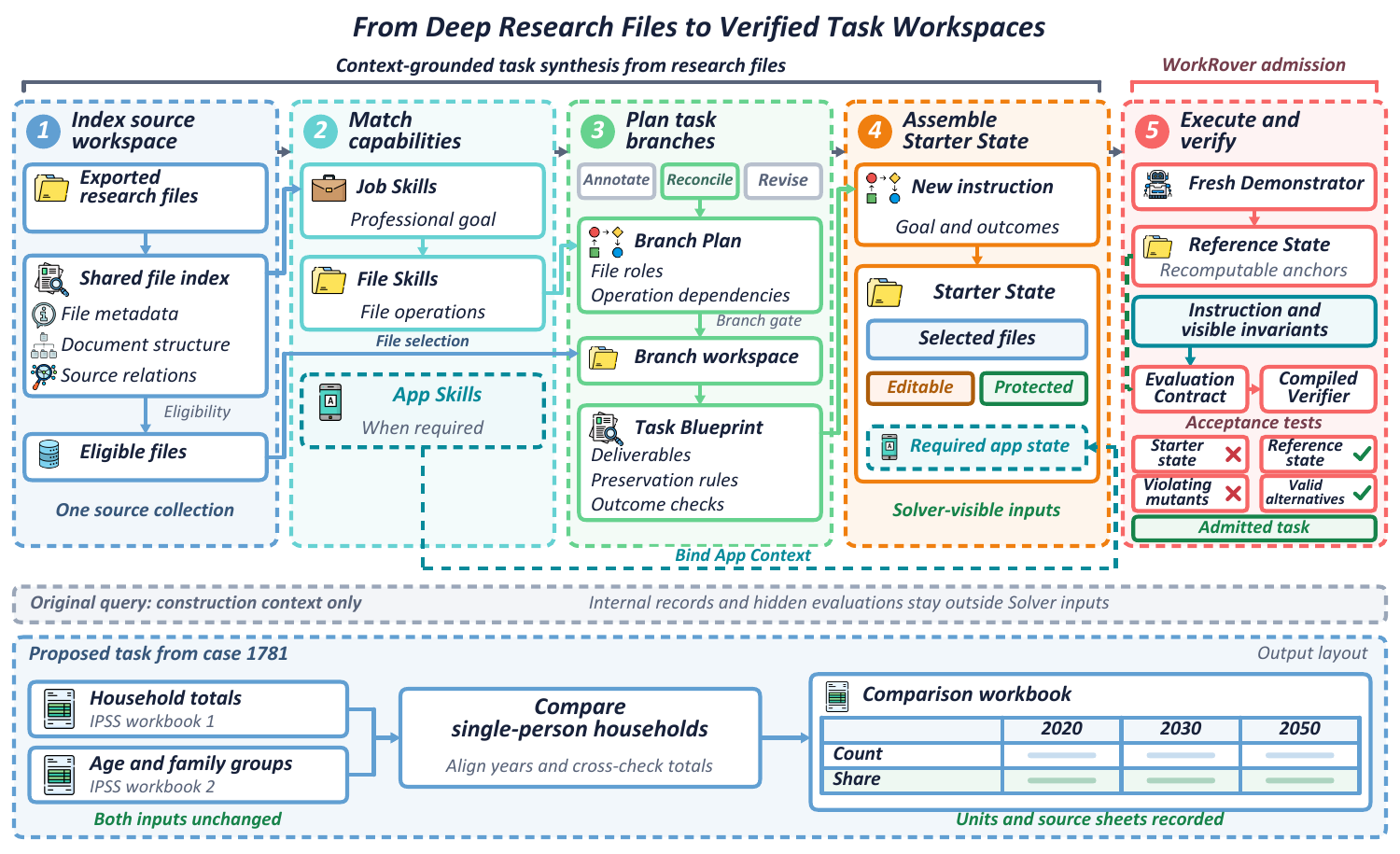}
\caption{\textbf{From Deep Research files to verified task workspaces.}
Instructions and visible invariants define requirements; the dashed green
arrow supplies recomputable Reference State anchors. Case 1781 uses two
IPSS workbooks; shaded cells mark output values to populate.}
\label{fig:deep-research-workspace-synthesis}
\end{figure}

In this route, a source workspace is the exported file collection associated
with one research record. Historical research supplies materials; new tasks
require fresh Demonstrator execution and independent admission
(Figure~\ref{fig:deep-research-workspace-synthesis}).

\paragraph{Inventory and source identity.}
Files are inventoried by relative path, format, size, and content hash.
Bounded previews expose headings, workbook sheets, or archive members for
selective reading. Captured sources and independently retained, readable
notes, tables, or reports may enter the Starter State after relevance and
leakage checks. Raw conversations, tool logs, internal analysis caches,
provenance directories, and unsupported web caches are excluded. The original
query informs construction only and is not Solver evidence. Source pointers
support audit; a URL cannot replace a required local file. Citations in notes
must be checked against their sources. Co-occurrence, retrieval order, and
model assertions do not establish dependencies; missing lineage is unresolved.
Freezing files does not make historical searches or services replayable.
Unavailable required sources lead to revision or rejection; permitted new
retrieval follows the Harness contract and records its evidence separately.

\paragraph{Branch plans and answer leakage.}
A job skill's requirements guide eligible-file retrieval within one source
collection. File skills bind the required operations; app skills bind any
instrumented application capabilities. Branches must differ in professional
decisions, operations, or principal products, and cannot import unselected
files or combine source collections. A feasible Branch Plan becomes a
workspace and Task Blueprint specifying file roles, dependencies, deliverables,
and modification/preservation constraints. Every required outcome has an
observable criterion. A write acknowledgment or readback excerpt alone is
not an editable artifact. Retained notes require evidence for their claims;
inputs that already complete the new task cause revision or rejection.
Hidden plans, solution keys, and evaluation records stay outside Solver
inputs and cannot supply unstated requirements or inaccessible task evidence.

\paragraph{Fresh execution, verification, and lineage.}
The instruction and selected files/application state define a frozen Starter
State. A fresh Demonstrator establishes feasibility and a Reference State;
historical session completion cannot substitute. The Contract uses only the
instruction and visible invariants, with independently recomputable anchors.
Admission requires rejection of the Starter State and violating mutants,
acceptance of valid Reference States and alternatives, and protected-state
checks. Branches retain collection, session, and file-version lineage.
Sessions, retained collections, candidates, and admitted tasks are counted
separately; branch count alone does not establish independent diversity.

\subsection{Task Synthesis and Quality Control}
\label{app:task-construction-protocol}
Each package contains an instruction, resettable snapshot, environment
specification, metadata, and a fresh Demonstrator realization with available
terminal artifacts. Instructions state outcomes and constraints without
prescribing tools or action sequences. Core files require defensible relations;
input ablation removes unnecessary bindings, and distractors cannot change
the dependency closure. Table~\ref{tab:task-construction-profiles} controls
scale and interaction horizon.

\begin{table}[!htbp]
    \centering
    \caption{Target task-construction profiles. The profiles control scale and
    interaction horizon.}
    \label{tab:task-construction-profiles}
    \small
    \setlength{\tabcolsep}{4pt}
    \TableOuterRules{%
\resizebox{0.96\columnwidth}{!}{%
    \begin{tabular}{lccc}

        \textbf{Dimension} & \textbf{P1} & \textbf{P2} & \textbf{P3} \\
        \hline
        Edit scope (files) & $\sim$10 & $\sim$20 & $>30$ \\
        Demonstrator horizon target (turns) & 5--10 & 15--30 & $>30$ \\
        Skill composition target & 2 job + $\geq1$ file & 3 job + $\geq3$ file & 4 job + $\geq5$ file \\
        App skills & As required & As required & As required \\
        WorkEnv noise ratio & 10\% & 20\% & 30\% \\
        Cross-env dependencies & 2 & 3 & 4 \\
        Distractor tiers & Obvious & Obvious + near-topic & All three \\

    \end{tabular}}
}
\end{table}

Empirical difficulty is the valid-rollout failure rate of a frozen reference
panel under matched snapshots, tools, permissions, and budgets. Panel
membership, aggregation, and band thresholds are fixed on development data;
evaluated systems do not contribute to these labels. Gates check instruction
clarity, readable inputs, sufficient dependencies, reproducible restoration,
and outcome anchors. Unobservable core outcomes are rejected. External
capabilities require declared Harness tools with recorded versions, seeds,
permissions, and initial-state identities.

\subsubsection{Task Design and Review}
\label{app:task-design-review}
The Blueprint binds each outcome to extractable or independently recomputable
artifact/application evidence (Table~\ref{tab:workflow-outcome-evidence}).
The manifest distinguishes required inputs, history, distractors, and protected
artifacts. Review checks opening/parsing, formulas, merged cells, annotations,
version history, layout, and embedded objects; visually specified outcomes
also require rendered inspection.

\begin{table}[!htbp]
\centering
\small
\setlength{\tabcolsep}{4pt}
\renewcommand{\arraystretch}{1.08}
\caption{Representative workflow requirements and observable outcome evidence.}
\label{tab:workflow-outcome-evidence}
\TableOuterRules{%
\begin{tabular}{@{}p{0.24\linewidth}p{0.34\linewidth}p{0.37\linewidth}@{}}
\textbf{Workflow} & \textbf{Inputs and dependencies} & \textbf{Outcome evidence} \\
\hline
Artifact revision & Source versions, comments, approvals & Required edits, consistent versions, preserved review history/protected content \\
Data reconciliation & Datasets, formulas, source records, periods & Recomputed aggregates, formulas, units, declared missing-value handling \\
Structured content & Reports, templates, charts, media & Required content/layout and preserved unaffected elements \\
Cross-artifact delivery & Linked files, output specifications, exports & Complete outputs, formats/order, source--export consistency \\
Collaboration & Calendar, email, drive, approval state & Required state changes, consistent records, satisfied workflow conditions \\
\end{tabular}}
\end{table}

\subsection{Case Studies: Recorded Deep Research Examples}
\label{app:deep-research-examples}

Figures~\ref{fig:deep-research-teaching-example}
and~\ref{fig:deep-research-policy-example} pair recorded research workflows
with their exported file collections. $m_i$ is the zero-based trace message index.
The proposed file tasks illustrate Appendix~\ref{app:deep-research-construction}.
Shaded cells indicate output fields to populate.

\begingroup
\setlength{\intextsep}{6pt}
\begin{figure}[!htbp]
    \centering
    \includegraphics[width=\linewidth]{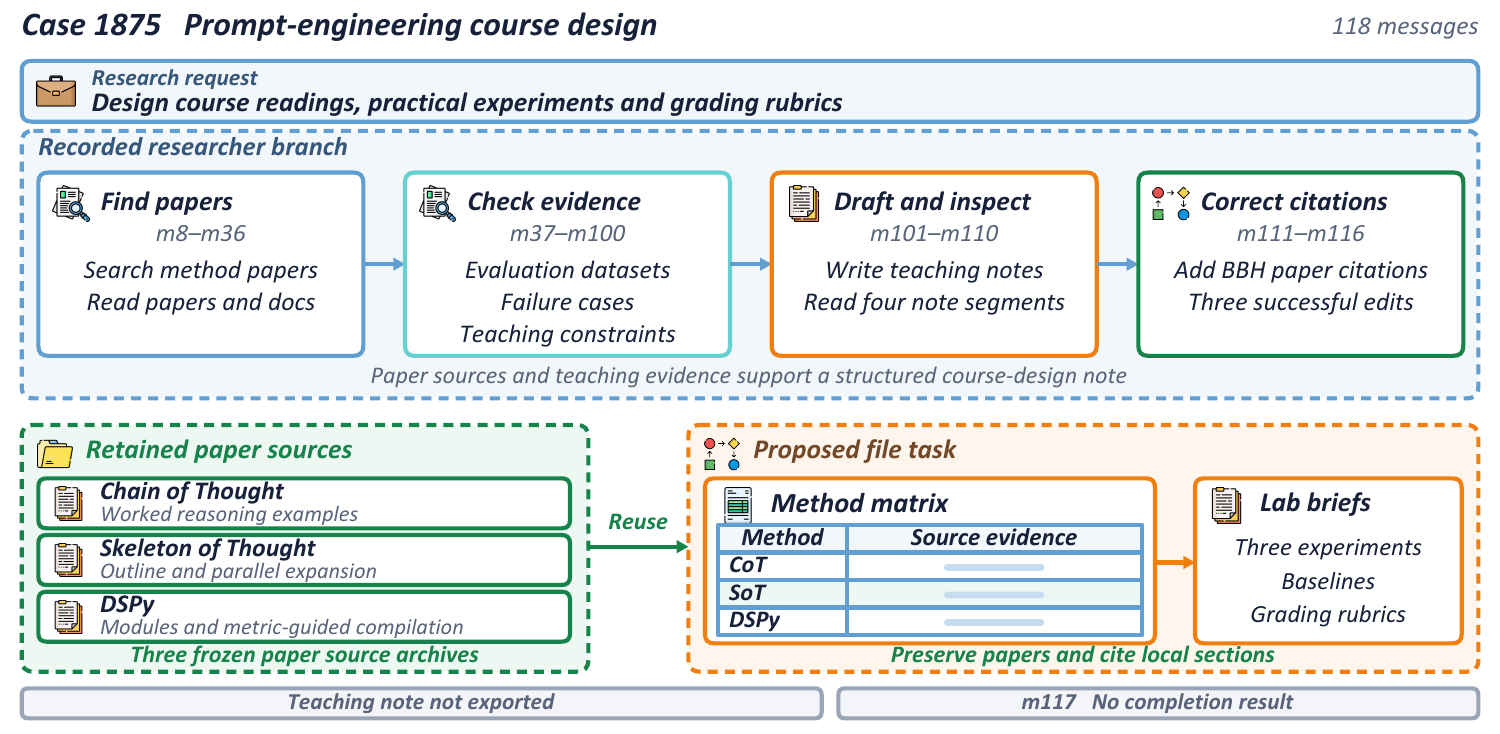}
    \caption{\textbf{Recorded example 1875: prompt-engineering course design.}
    This researcher branch writes \texttt{teaching\_design.md}, inspects four
    segments, and adds the BIG-Bench Hard paper citation in three places. The proposed task
    reuses captured CoT, SoT, and DSPy paper-source archives to produce a method
    comparison workbook and three lab briefs with baselines and grading rubrics.
    Claims cite local source sections; the original archives remain unchanged.}
    \label{fig:deep-research-teaching-example}
\end{figure}

\begin{figure}[!htbp]
    \centering
    \includegraphics[width=\linewidth]{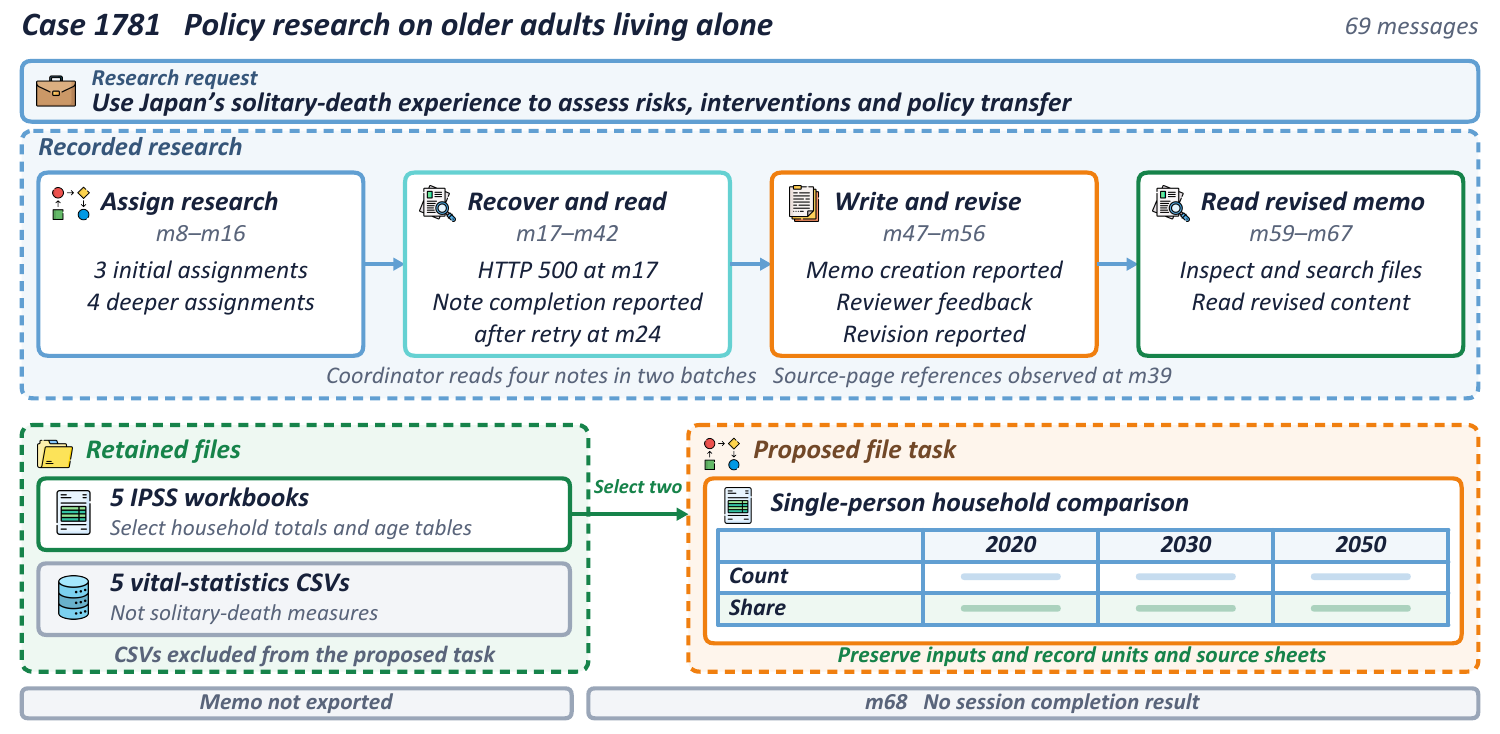}
    \caption{\textbf{Recorded example 1781: policy research on older adults living alone.}
    The proposed task selects \texttt{hprj2024\_gaiyo\_kekka1.xlsx} and
    \texttt{hprj2024\_gaiyo\_kekka2.xlsx}, retaining category definitions, units,
    and source-sheet references. The revised \texttt{policy\_memo.md} is
    observed through tool readbacks. Source-page references at $m_{39}$ cite
    IPSS and MHLW.}
    \label{fig:deep-research-policy-example}
\end{figure}
\endgroup

Source-page references associate these files with the research but do not
establish per-file downloads in the recorded traces. New tasks
still require fresh Demonstrator execution and counterfactual Verifier checks;
raw traces and internal provenance remain outside Solver inputs.

\subsection{Case Study: Quarterly Executive Handoff}
\label{app:rl-task-example}
Record \texttt{0805:comp\_018554} belongs to the 5,000-trajectory pool and
the 3,200-trace subset described in Appendix~\ref{app:experimental-setup}.
Figure~\ref{fig:rl-task-example} follows its Hermes execution.
The agent verifies five regional revenue rows, registers the vendor, amends
the Q2 access policy, retires the archived Q1 policy, and checks five exports.
Revenue values match the source; the workbook adds totals, growth, and evidence
references. The final report retains the database lock wait and stale export
alias as follow-ups. All four archived goal groups pass (score 0.946);
file validation covers structure/content, with visual review unavailable.

\begin{figure}[!htbp]
\centering
\includegraphics[width=0.94\linewidth]{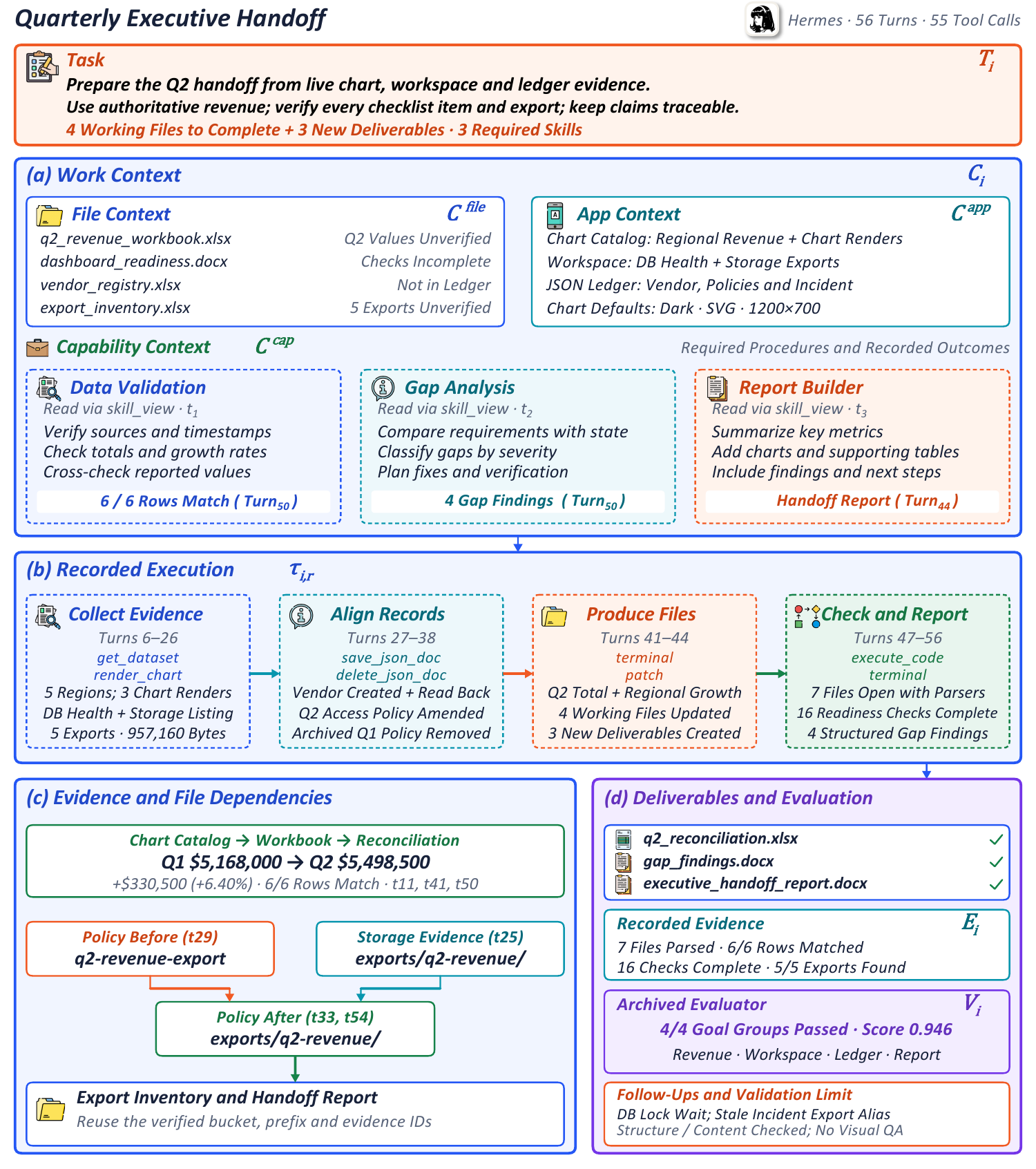}
\caption{\textbf{Recorded quarterly executive handoff.}
(a) File, capability, and app context. (b) Selected rollout stages.
(c) Revenue and export-path evidence supporting consistent updates.
(d) Deliverables, archived evaluation, retained follow-ups, and validation scope.}
\label{fig:rl-task-example}
\end{figure}

\section{Verifier, Repair, and Training Details}

\subsection{Verifier Specification and Calibration}
\label{app:verifier-specification}
Criteria derive from the instruction, Starter State, and visible invariants;
Reference artifacts only instantiate independently recomputable anchors.
The frozen Contract specifies editable/protected state, allowed transitions,
and each criterion's predicate, evidence, checking mode, weight $w_k$,
threshold $\gamma_k$, and critical flag $b_k$. Unresolved requirements prevent
admission. Adapters and canonical diffs provide stable evidence IDs. Executable
checks read both snapshots; at most one batched Judge call receives frozen
criterion-tagged evidence, has no live tools, and cites evidence or abstains.
For scores $p_{i,k}\in[0,1]$, set
$y_{i,k}=\mathbf1[p_{i,k}\geq\gamma_{i,k}]$ and
$\bar w_{i,k}=w_{i,k}/\sum_jw_{i,j}$:
\begin{equation}
 S_{B,i}=\sum_k\bar w_{i,k}p_{i,k},\qquad
 H_i=g_i^{\mathrm{integrity}}g_i^{\mathrm{protected}}g_i^{\mathrm{safety}},
 \qquad G_i=H_i\prod_{k:b_{i,k}=1}y_{i,k}.
 \label{eq:outcome-score-gates}
\end{equation}
Healthy evaluation passes iff $G_i=1$ and $S_{B,i}\geq\tau_B$.
\label{eq:three-valued-decision}
Missing required state implies failure; infrastructure-unavailable evidence
implies invalidity. Policy errors under healthy infrastructure are not invalid.
Partial credit cannot offset gates. Auxiliary $S_A\in[0,1]$ uses frozen trace
checks. Admission requires the Starter--Reference contrast, protected-state
checks, no unresolved invalid controls, and frozen thresholds for criterion
agreement, worst-family mutant detection, and valid-alternative acceptance.
Task-level negatives must violate the Contract. Calibration, initial admission,
reporting, and promotion use disjoint recipe--seed banks. Unsupported semantic
criteria retain terminal supervision without local progress signals.

\subsection{FCR Mechanics and Promotion}
\label{app:fcr-mechanics}
Freeze one direct proposal per available Context/Task/Harness/Verifier factor
before replay: at most 16 cells, or eight before compilation. Each cell $\chi$
applies the dependency-closed update set $\Gamma(\chi)$, including
dependency-induced updates; changed tasks require fresh execution/compilation,
and shared Harness edits require affected-task checks. Established conflicts
make cells unavailable; uncertain cells proceed to materialization.
Candidate repair returns to the earliest affected gate; maintenance freezes
the admitted parent. Both preserve intent, outcomes, transitions, and applicable
Contract weights, thresholds, flags, and gates. Hidden evidence cannot leak
into visible edits; new tools require separate releases and semantic changes
create new candidates. Blinded equivalence review and a frozen solver panel
apply development-fixed simplification margins.

Matched Context--Task--Harness seeds produce evidence reused by Verifier
variants. Health uses all seeds; outcome contrasts use common healthy seeds
on complete compatible subcubes, without imputation. Verifier contrasts use
sealed label agreement. Selection requires a proposal-independent defect/health
predicate, proposal-disjoint replay, equivalence, no affected-task regression,
and compiled-Verifier calibration plus a one-use binary promotion shard.
Reporting audits never select repairs. The smallest validated closed edit set $\Gamma(\chi^\star)$
among tested cells receives fresh-seed confirmation
(Section~\ref{sec:workspace-coevolution}); failure discards candidates or
retains maintenance parents and quarantines unresolved triggers.

\subsection{FCR Evaluation Protocol}
\label{app:fcr-evaluation}
Frozen stratified screening of the 10K pool targets 300 natural mismatches,
classified by two blinded reviewers with third-reviewer adjudication.
The separate controlled cohort injects 260 concealed Contract-preserving
defects: 80 single-, 120 pair-, 40 triple-, and 20 four-factor cases.
Candidate-controlled comparisons share proposals; end-to-end comparisons allow
independent proposals under matched caps. Each distinct Context--Task--Harness
configuration uses five matched seeds; compatible Verifiers reuse evidence.
Confirmation seeds and generation-specific promotion shards are disjoint.
Joint necessity excludes passing strict proposal subsets. Triggers are
statistical units; seeds/cells are nested measurements.

\subsection{Verifier-to-Turn Credit Assignment}
\label{app:vtca-details}
Canonical traces retain ordering, call/result IDs, tool definitions, versions,
and state metadata. SFT supervises agent tokens; invalid rollouts receive no
training signal. For valid rollout $r$,
\begin{equation}
 R_{i,r}=G_{i,r}[(1-\lambda)S_{B,i,r}+\lambda S_{A,i,r}],\quad
 A^{\rm traj}_{i,r}=\frac{R_{i,r}-\bar R_i}
 {\max(\operatorname{sd}_{r'}R_{i,r'},\sigma_{R,\min})},\quad 0\leq\lambda\leq0.1.
 \label{eq:training-reward}
\end{equation}
The denominator floor is development-fixed. Deterministic progress
$\Delta_{i,r,t,k}=\phi_{i,k}(x_{i,r,t+1})-\phi_{i,k}(x_{i,r,t})$ requires
both endpoints evaluable; otherwise record zero increment and a missing reason.
Missing intervals are not bridged or treated as zero completion.
No intermediate Judge is called. Prerequisite set $\mathcal P_{i,r,s,k}$
deduplicates earlier producer turns whose exact outputs/versions are directly
consumed at $s$, with criterion-compatible immutable IDs and call/result links.
Stale, ambiguous, inferred, and unrecorded uses are excluded. With
$\rho_{i,r,s,k}=\rho\mathbf1[|\mathcal P_{i,r,s,k}|>0]$ and $0\leq\rho\leq1/2$,
\begin{equation}
\begin{aligned}
 b_{i,r,s,t,k}&=\begin{cases}
 1-\rho_{i,r,s,k},&t=s,\\
 \rho_{i,r,s,k}/|\mathcal P_{i,r,s,k}|,&t\in\mathcal P_{i,r,s,k},\\
 0,&\text{otherwise},
 \end{cases}\\
 \delta_{i,r,t}^{\mathrm{crit}}&=\sum_k\bar w_{i,k}\left[
 \sum_{s=t}^{T_{i,r}-1}b_{i,r,s,t,k}\max(\Delta_{i,r,s,k},0)
 +\min(\Delta_{i,r,t,k},0)\right].
\end{aligned}
\label{eq:vtca-propagation}
\end{equation}
Allocation is direct; negatives and empty-set increments stay local.
Weights are not renormalized over observed criteria or centered across turns.
Equation~\ref{eq:vtca-turn-advantage} gates progress by $H$, terminal reward
by $G$. Trace cost $c_t\in[0,1]$ covers malformed/error calls, redundant
state-preserving actions, and attributable protected-state/safety violations.
GRPO averages tokens within turns, turns within trajectories, then trajectories
within task groups, using fixed behavior policy, SFT reference, and detached
advantages. Before gating/cost/clipping,
$\sum_t\delta_{i,r,t}^{\mathrm{crit}}=\sum_k\bar w_{i,k}\sum_t\Delta_{i,r,t,k}$;
only complete observations telescope. Write--revert cycles and wrong
provenance can still alter updates.

\section{Evaluation Details Referenced in the Main Text}
\label{app:experimental-details}

\subsection{Experimental Setup}
\label{app:experimental-setup}
\paragraph{Training and visualization data.}
Qwen3.5-35B-A3B is the post-training backbone; 397B-A17B supplies 30,115
Verifier-accepted SFT trajectories passing structural/lineage QA. Collection
allows three retries per rejected turn with exact prefix restoration. Four
35B-A3B pilots per candidate retain 3,200 RL pairs from 20,000 candidates;
splits are by Work Context. RL starts from SFT with turn-balanced GRPO,
KL coefficient 0.01, and matched tools, permissions, states, and budgets.
Trajectory analysis separately selects 3,200 of
5,000 successful traces: $40<$ deduplicated assistant turns $<100$,
\texttt{eval.score}$>0.8$, passed evaluation, complete segments, and normal
termination. Ranking uses named skill calls, score, then deterministic hash.
This subset does not estimate online RL; valid failures still train. The Harness
allocation is OpenCode/Claude Code/Codex/Hermes 60/20/10/10\%.

\paragraph{Controlled evaluations.}
Matched runs fix release, split, evaluator, Harness, and budget, charging
invalid/retried attempts under frozen caps. Tasks cluster by Work Context.
GDPval Elo and percentages are not averaged. Controlled OfficeQA Pro uses
133 questions at 0.0\% numerical tolerance.

\paragraph{Benchmark configuration.}
GDPval v2 uses the September 12, 2026 (Asia/Shanghai) AA snapshot and Stirrup
\citep{artificialanalysis2026gdpvalaa}.
JobBench Main uses 65 tasks, OpenCode v1.14.18, and Grok 4.3
\citep{jobbench2026leaderboard}.
For OfficeQA Pro, \textsc{WorkRover} uses Claude Code, temperature
0.7, top-$p$ 0.95, top-$k$ 40, 1,800 seconds.

\subsection{Matched VTCA Comparison}
\label{app:training-results}
Figure~\ref{fig:vtca-main-results} matches checkpoints, task/seed schedules,
groups, rollout/token/update budgets, and interaction/Verifier/Judge caps;
all attempts are charged. VTCA uses one terminal Verifier pass and at most
one Judge call; checker/capture/CPU costs are separate. Controls cover
aggregation, local-only progress ($\rho=\eta=0$), matched random/verified
recipients, and trace cost. Random recipients preserve producer mass, budget,
and count. Inference resamples training replicas and task/Context clusters;
GDPval refits ratings from matched judgments.

\section{Data Construction and Case Study Summaries}
\label{app:construction-case-summaries}
The following tables consolidate the material boundaries and admission process
in Appendices~\ref{app:deep-research-construction} and~\ref{app:task-construction-protocol},
and distinguish proposed tasks from the recorded execution in
Appendices~\ref{app:deep-research-examples} and~\ref{app:rl-task-example}.

\begin{table}[!htbp]
\centering
\small
\setlength{\tabcolsep}{4pt}
\renewcommand{\arraystretch}{1.08}
\caption{Source materials, construction roles, and Solver visibility.
Eligibility does not establish factual authority or evidence sufficiency.}
\label{tab:source-material-visibility}
\TableOuterRules{%
\begin{tabular}{@{}>{\RaggedRight\arraybackslash}p{0.21\linewidth}>{\RaggedRight\arraybackslash}p{0.36\linewidth}>{\RaggedRight\arraybackslash}p{0.39\linewidth}@{}}
\textbf{Material} & \textbf{Construction role} & \textbf{Solver visibility} \\
\hline
Captured source files &
Readable source documents, code, or tables. &
Selected files are frozen in the Starter State. \\
Retained working files &
Editable files; factual claims require supporting sources. &
May be selected if they do not reveal the completed answer. \\
Original research query &
Background for job skills and task intent. &
Construction only; never Solver input or evidence. \\
Logs and internal analysis &
Audit context; readbacks are not editable artifacts. &
Excluded from selectable inputs. \\
Provenance and source pointers &
Lineage/source identity; URLs cannot replace local files. &
Internal manifests remain outside Solver inputs. \\
\end{tabular}}
\end{table}

\begin{table}[!htbp]
\centering
\small
\setlength{\tabcolsep}{4pt}
\renewcommand{\arraystretch}{1.08}
\caption{Task construction stages and admission checks. Each new task requires
fresh execution and verification, independently of historical session completion.}
\label{tab:task-construction-gates}
\TableOuterRules{%
\begin{tabular}{@{}>{\RaggedRight\arraybackslash}p{0.19\linewidth}>{\RaggedRight\arraybackslash}p{0.37\linewidth}>{\RaggedRight\arraybackslash}p{0.40\linewidth}@{}}
\textbf{Stage} & \textbf{Recorded output} & \textbf{Acceptance check} \\
\hline
Index the source &
Paths, formats, sizes, hashes, structural previews, and available file relations. &
Selected inputs are readable; claimed dependencies have support and unresolved lineage is recorded. \\
Plan a task branch &
Job/file/app skill bindings and a Branch Plan selecting files from one collection. &
The objective is feasible; branches differ in decisions, operations, or products; no unselected files are imported. \\
Assemble the Starter State &
Frozen files and app state, permissions, capability bindings, and a role/protection manifest. &
Dependencies are available; reset/recovery works; task-relevant structures survive; no completed-answer leakage. \\
Specify and execute &
Instruction, Task Blueprint, fresh Demonstrator trace, and Reference State. &
Required outcomes and preservation constraints are observable and achievable under the declared Harness. \\
Compile and calibrate &
Frozen Contract, recomputable anchors, evidence bindings, and counterfactual controls. &
The Starter State fails; valid Reference States and alternatives pass; violating mutants and protected-state changes are detected. \\
Record lineage and split &
Source/session/file-version identities and separate collection, candidate, and admission counts. &
Connected lineage and derived branches remain in one split; branch count is not treated as independent diversity. \\
\end{tabular}}
\end{table}

\begin{table}[!htbp]
\centering
\small
\setlength{\tabcolsep}{4pt}
\renewcommand{\arraystretch}{1.08}
\caption{Case study inputs, work products, and evidence scope. The two Deep
Research cases illustrate proposed derivative tasks; the handoff case is a recorded execution.}
\label{tab:case-study-summary}
\TableOuterRules{%
\begin{tabular}{@{}>{\RaggedRight\arraybackslash}p{0.19\linewidth}>{\RaggedRight\arraybackslash}p{0.40\linewidth}>{\RaggedRight\arraybackslash}p{0.37\linewidth}@{}}
\textbf{Case} & \textbf{Inputs and requested work products} & \textbf{Evidence and verification scope} \\
\hline
1875: course design &
Captured CoT, SoT, and DSPy paper archives support a method comparison workbook and three lab briefs with baselines and grading rubrics. &
Claims cite local source sections; archives remain unchanged. The historical teaching note was not exported. Fresh task execution and admission are still required. \\
1781: household projections &
Two IPSS workbooks support a comparison of single-person household counts/shares for 2020, 2030, and 2050, retaining definitions, units, and source-sheet references. &
Both input workbooks remain unchanged. Historical memo evidence is limited to tool readbacks; source-page references do not prove individual downloads. Fresh admission is required. \\
Quarterly executive handoff &
Revenue sources, working files, chart/workspace state, and the ledger support updated files, a reconciliation workbook, gap findings, and an executive handoff report. &
Record \texttt{0805:comp\_018554}: all four goal groups passed, score 0.946. Structure/content were checked; visual review was unavailable. Lock-wait and stale-alias follow-ups remain. \\
\end{tabular}}
\end{table}

\end{document}